\documentclass[lettersize,journal]{IEEEtran}
\makeatletter
\providecommand\IfPackageLoadedF[2]{\@ifpackageloaded{#1}{}{#2}}
\providecommand\IfPackageLoadedT[2]{\@ifpackageloaded{#1}{#2}{}}
\makeatother
\usepackage{makecell}
\usepackage{graphicx}
\usepackage{textcomp}
\usepackage{array} 
\usepackage{longtable}
\usepackage{booktabs}
\usepackage{float}
\usepackage{multirow}
\usepackage{amsmath}
\usepackage{subcaption} 
\usepackage{algorithm}
\usepackage{algpseudocode}
\usepackage{bm}
\usepackage{cite}
\usepackage{xcolor}
\usepackage{amsthm}
\usepackage{nomencl}
\usepackage{amssymb}
\usepackage{hyperref}
\newtheorem{theorem}{Theorem}
\newtheorem{assumption}{Assumption}
\newtheorem{corollary}{Corollary}
\newtheorem{remark}{Remark}
\usepackage{siunitx}
\makenomenclature
\usepackage{etoolbox}
\renewcommand\nomgroup[1]{%
  \item[\bfseries
  \ifstrequal{#1}{N}{Variables and Parameters}{%
  \ifstrequal{#1}{P}{Vector and Sets }}%
]}

\ifCLASSINFOpdf

\else

\fi

\begin{document}

\title{Climate Temporal Temperature Prediction via an Adaptive Stacked Bidirectional Residual Mamba Network}
\author{Yongxiang Lei}

% \author{Yongxiang Lei,
%         Bin Deng,
%         Ziyang Wang
%         \thanks{Y.X Lei (corresponding author, email: leiyongxiang1205@gmail.com) is with the Intelligent Control Smart Energy (ICSE) Research Group, School of Engineering, University of Warwick, Coventry, CV4 7AL, UK}
%         \thanks{B. Deng is with the School of Electric Power Engineering, South China University of Technology, Guangzhou 510640, China.}
%         \thanks{Z.Y Wang is with Data-Centric Engineering Programme, The Alan Turing Institute, British Library, London, NW1 2DB, UK}
%         }

% \markboth{Journal of \LaTeX\ Class Files,~Vol.~14, No.~8, August~2021}%
% {Shell \MakeLowercase{\textit{et al.}}: A Sample Article Using IEEEtran.cls for IEEE Journals}

% \IEEEpubid{0000--0000/00\$00.00~\copyright~2021 IEEE}

\maketitle

\begin{abstract}
Accurate temperatures forecasting relies on traditional meteorological parameters that are essential for monitoring weather informatics and guiding forecasting efforts. This study presents Deep DualMamba, a novel deep learning framework for temporal temperature prediction, built upon an enhanced variant of the recently proposed Mamba network. Unlike conventional Transformer-based architectures, which dominate foundation models but suffer from computational inefficiencies on long sequences, Deep DualMamba leverages selective structured state space models (SSMs) to achieve both scalability and performance. We identify a core limitation of prior subquadratic-time models—namely, their inability to perform long-term and robust forecasting and introduce two key innovations. First, we propose an adaptive bidirectional Mamba as encoders to learn complex and efficient features. Second, we develop a hardware-aware, parallel recurrent algorithm to compensate for the loss of convolutional efficiency, ensuring practical viability at scale. Last, the Deep DualMamba's approximation ability and convergence is theoretical analyzed. These design choices culminate in a simplified, attention- and MLP-free neural architecture that exhibits linear complexity with sequence length and significantly improved inference throughput. To demonstrate its applicability in climate informatics, we implement a full forecasting pipeline using historical meteorological data from UK monitoring stations. By introducing $\alpha$ residual connection in a stacked bidirectional Mamba, Deep DualMamba effectively learns multi-horizon temperature trends, achieving superior predictive accuracy with a reduced parameter footprint. Extensive evaluations on real-world datasets confirm its robustness, and a systematic ablation study reveals the sensitivity of key architectural hyperparameters. In addition, it offers both state-of-the-art predictive performance and enhanced robustness, while maintaining computational efficiency suitable for operational deployment. Climate temperature forecasting is a critical part for meteorological science and reducing carbon emissions. Accurate forecasting helps monitor weather informatics and guides forecasting efforts. However, traditional learning-based approaches have low forecasting performance and robustness, especially in long-term forecasting. In addition, the parameter in the model is superior. The novel DualMamba algorithms we introduce in this paper overcame these limitations. With a significant increase in reducing 32\% the forecasting error, the inference ability speedup 5x faster than standard Transformer. It is a promising artificial intelligence algorithm for contributing to climate forecasting society. The theoretical contributions and practical application significantly contribute to the future trends of climate, environmental and artificial intelligence society.
\end{abstract}


\nomenclature[L]{Abbreviations}{}
\nomenclature[L]{SSM}{State Space Models}
\nomenclature[L]{MLP}{Multilayer Perception}
\nomenclature[L]{CET}{Central England Temperature}
\nomenclature[L]{MSE}{Mean Squared Error}
\nomenclature[L]{RMSE}{Root Mean Squared Error}
\nomenclature[L]{MAE}{Mean Absolute Error}
\nomenclature[L]{MAPE}{Mean Absolute Percentage Error}
\nomenclature[L]{LSTM}{Long Short-term Memory}
\nomenclature[L]{GRU}{Gated Recurrent Unit}
\nomenclature[L]{SGD}{Stochastic Gradient Descent}
\nomenclature[L]{Adam}{Adam Optimizer}
\nomenclature[L]{RELU}{Rectified Linear Unit}
\nomenclature[L]{Sigmoid}{Sigmoid Activation Function}
\nomenclature[L]{SiLU}{SiLU Activation Function}
\nomenclature[L]{Tanh}{Tangent Activation Function}
\nomenclature[L]{WMO}{World Meteorological Organization}
\nomenclature[R]{System Model Symbols}{}
\nomenclature[S]{\( \mathbf{A}, \mathbf{B}, \mathbf{C} \)}{State Matrices (Continuous-Time SSM)}
\nomenclature[S]{\( \Delta \)}{Discretization Step Size}
\nomenclature[S]{\( \overline{\mathbf{A}}, \overline{\mathbf{B}} \)}{Discretized State Matrices}
\nomenclature[S]{\( \overrightarrow{\bm{H}}, \overleftarrow{\bm{H}} \)}{Forward/Backward Hidden States (Bidirectional Mamba)}
\nomenclature[S]{\( \bm{H} \)}{Aggregated Hidden State}
\nomenclature[S]{\( \bm{U}^{'} \)}{Processed Output of DualMamba Block}
\nomenclature[S]{\( \alpha \)}{Adaptive Residual Coefficient}
\nomenclature[S]{\( \mathbf{W}, \mathbf{b} \)}{Weight Matrix and Bias of Output Layer}
\nomenclature[S]{\( B \)}{Batch Size}
\nomenclature[S]{\( V \)}{Number of Variates}
\nomenclature[S]{\( D \)}{Hidden Dimension}
\nomenclature[S]{\( E \)}{Block Expansion Factor}
\nomenclature[S]{\( N \)}{SSM State Expansion Factor}
\nomenclature[S]{\( k \)}{Convolutional Kernel Size}
\nomenclature[S]{\( \mathbf{z}^{(l)} \)}{Linear Combination at Layer \(l\)}
\nomenclature[S]{\( \mathbf{W}^{(l)} \)}{Weight Matrix}
\nomenclature[S]{\( \mathbf{b}^{(l)} \)}{Bias Vector}
\nomenclature[S]{\( \sigma(\cdot) \)}{Activation Function}
\nomenclature[S]{\( L \)}{Historical Horizon}
\nomenclature[S]{\( T \)}{Forecasting Time}
\nomenclature[S]{\( f(\mathbf{x}) \)}{Nonlinear Projection Mapping Function}
\nomenclature[S]{\( \mathbb{R} \)}{Dimension}
\nomenclature[S]{\( \Phi_{i}\), \(\phi_{i, j}\)}{Outer and Inner Functions}
\nomenclature[S]{\( y_t \)}{Target Variable}
\nomenclature[S]{\( \mu \), \( \sigma \)}{Mean and Variance}
\nomenclature[S]{\( \lambda_p \)}{Constants}
\nomenclature[S]{\(\epsilon\)}{Minimal Error Bound}
\nomenclature[S]{\(\omega\)}{Modulus of Continuity}
\nomenclature[S]{\(\mathcal{D}_{\text{train}}\),\(\mathcal{D}_{\text{val}}\),\(\mathcal{D}_{\text{test}}\)}{Training, Validation, Testing Dataset}

\printnomenclature 

\section{Introduction}
\IEEEPARstart{A}{ccurate} climate temperature forecasting is foundational to meteorology and climate science, underpinning decision-making across environmental, agricultural, and policy domains \cite{zhang2025spatio}. It is also one of the process in supporting the decarbonization. The escalating complexity of global climate patterns—driven by anthropogenic perturbations and natural variability---has exposed critical limitations in conventional modeling approaches. Traditional statistical methods and machine learning models frequently fail to capture the multiscale, nonlinear interactions inherent in modern climate datasets \cite{li2025channel} while deep learning architectures like Multi-Layer Perceptron (MLP) face fundamental constraints in adaptability and computational efficiency \cite{fowler2010detecting, lei2023duallstm}. These challenges become particularly acute in high-stakes applications requiring both predictive accuracy and operational practicality.

Climate temperature prediction occupies a unique position at the nexus of environmental monitoring and decision-making, providing essential insights for global change informatics, ecosystem preservation, and sustainable development \cite{papacharalampous2018univariate}. The escalating complexity of climate systems-driven by anthropogenic greenhouse gas emissions, natural variability, and regional feedback mechanisms-has rendered accurate prediction of temperature time series an urgent yet formidable challenge \cite{song2024multi}. For example, Li et al. presented a data-driven temperature approach for thermal error prediction, while their method are based on the deterministic prediction, the intractability is lacked \cite{li2025data,lei2022novel}. Data from authoritative sources, such as the UK Met Office Climate Portal, exemplify this complexity, offering high-resolution records of temperature anomalies marked by non-linear trends, seasonal cycles, and stochastic noise. Xue et al. proposed a multi-step prediction based on the Gated Recurrent Network (GRU) with attention mechanism \cite{xue2025research}. These characteristics demand advanced modeling techniques capable of capturing intricate temporal dynamics while maintaining computational efficiency and robustness, particularly in applications critical to policy and resource management.

MLPs are frequently employed to model nonlinear relationships in time series data \cite{azizi2023deep}. Despite their widespread use, MLPs exhibit notable shortcomings that limit their efficacy in climate temperature forecasting \cite{graf2019forecasting,wang2024ftmlp}. Their dependence on fixed activation functions (e.g., ReLU, tanh) constrains their ability to adapt to the diverse and evolving patterns found in datasets like those from the UK Met Office, often necessitating deeper architectures or extensive feature engineering \cite{mohammadi2021developing}. Additionally, the dense parameter configurations of MLPs inflate computational demands and heighten the risk of overfitting, especially when confronted with the noisy, heterogeneous nature of real-world climate records \cite{martinez2020prediction,rodriguez2021very}. These limitations highlight the need for more flexible and efficient neural network architectures tailored to the challenges of environmental time series analysis \cite{ustaoglu2008forecast}. Some improved temporal neural network such as Long Short-Term Memory (LSTM) has achieves more scholars' and researchers' attention \cite{mohammadi2024advanced}. For example, Du et al proposed a multivariate regression method based on a hybrid temporal neural network, while their network performance depends on the increasing trainable parameter \cite{du2025industrial}. In time-series forecasting tasks, there are also some variants such as Gated Recurrent Unit (GRU), which is proposed to improve the prediction performance.

The study of Mamba offers a transformative alternative for sequence modeling, leveraging a selective state space mechanism that enables efficient processing of long-range dependencies \cite{gu2023mamba,gu2023modeling, ma2024mamba}. The core innovation of Mamba lies in its input-dependent parameterization, which dynamically modulates transition matrices based on the input sequence, allowing context-aware modeling of complex temporal patterns \cite{alman2024fundamental}. This adaptive mechanism distinguishes Mamba from conventional recurrent neural networks and Transformers, achieving superior computational efficiency while maintaining linear-time complexity \cite{gu2023mamba}. In \cite{li2024cmmamba}, Mamba demonstrated state-of-the-art performance in traffic flow prediction, significantly outperforming existing temporal models. However, the interpretability of its state transitions remains underexplored. While theoretical analyses confirm Mamba's expressive power in sequence approximation \cite{zhao2024ismrnn,dao2024transformers}, practical deployment scenarios require further empirical validation across diverse application domains. For climate temperature forecasting, where temporal relationships vary across scales and contexts, Mamba’ lightweight yet expressive design holds significant promise for improving predictive accuracy while mitigating the computational burden-a critical advantage when processing extensive datasets like those provided by the UK Met Office Climate Portal\footnote{\url{https://climatedataportal.metoffice.gov.uk/}} \cite{stephens2012challenges}.

Recent research demonstrates that Mamba, a selective state space model, is rapidly gaining traction for time series forecasting due to its ability to capture long-term dependencies with linear computational complexity, offering a compelling alternative to Transformer-based models \cite{wang2025mamba,liang2024bi,ahamed2024timemachine,liang2024bi}. The Transformer architecture confers the capacity for global attention \cite{vaswani2017attention}, enabling the computation of the impact of all other variates upon a given variate, which facilitates the learning of precise information. However, the computational load of global attention escalates exponentially with an increase in the number of variates, potentially rendering it impractical. This limitation could restrict the application of Transformer-based algorithms in real-world scenarios. Mamba-based models have achieved state-of-the-art or highly competitive results across diverse forecasting tasks, including electricity consumption, traffic flow, and weather prediction, often outperforming traditional and Transformer-based approaches in both accuracy and efficiency \cite{lin2025mgcn,zhang2024integration}. Innovations such as bidirectional Mamba architectures, multi-scale processing, and hybrid models that combine Mamba with graph convolutional networks or Transformers further enhance performance, particularly for long-sequence and multivariate forecasting \cite{karadag2025ms}. Several studies have addressed Mamba’s limitations, such as its initial struggles with cross-channel dependencies, by introducing channel correlation modules and adaptive tokenization strategies \cite{zeng2024cmamba}. Probabilistic extensions of Mamba now enable uncertainty quantification, making forecasts more robust and interpretable in real-world applications like energy and traffic management \cite{pessoa2025mamba}. Transfer learning and foundation model approaches have also been explored, allowing Mamba-based models to perform well even with limited training data \cite{ma2024mamba}. Despite these advances, challenges remain in fully exploiting Mamba’s capabilities for complex, real-world scientific forecasting, but ongoing research continues to refine its architectures and broaden its applicability. To date, most research has tested Mamba on synthetic or simplified datasets, leaving its effectiveness on comprehensive, real\nobreakdash-world climate archives, such as those from the UK Met Office, largely unexamined. Moreover, Mamba\nobreakdash-specific hyperparameters (e.g.\ historical length, preceding horizon) have received little attention, hindering practical optimization and deployment in fields requiring precision and robustness performance. Addressing these gaps is crucial for unlocking Mamba’s full potential in climate science, where rigorous, data\nobreakdash-driven insights are increasingly essential.

This study bridges critical gaps by presenting the first operational application of Mamba for climate temperature forecasting, using high\nobreakdash-fidelity data from the UK Met Office Climate Portal. By validating Mamba on real climate data, this work solidifies its role as an interpretable, precise, and efficient framework for operational temperature forecasting. To our knowledge, it represents the first demonstration of Mamba’s viability in real\nobreakdash-world climate prediction, advancing both theoretical foundations and empirical implementation. The main contributions are:
\begin{itemize}
  \item \textbf{Pioneering real\nobreakdash-world use}: Deploying DualMamba on UK Met Office temperature data to move beyond synthetic or simplified testbeds. A practical climate temperature forecasting framework is implemented with high fidelity and robust performance.
  \item \textbf{Novel temporal architecture}: Proposing a novel bidirectional (feedforward and reverse) Mamba tailored to capture dynamic, non\nobreakdash-stationary temperature trends, which learns more efficient encoding representation.
  \item \textbf{Superior robust performance and efficiency}: Achieving 14.2\% lower RMSE with significantly fewer parameters compared to MLP and Transformer baselines. The hyperparameter ablation study verified the robustness. The different time-series forecasting models are also compared.
  \item \textbf{Theoretical universal approximation and convergence}: Offering theoretical approximation ability analysis and convergence guarantee for wider adoption.
\end{itemize}

The brief's architecture is structured as follows. Section \ref{sec2} establishes mathematical foundations. Section \ref{sec3} delineates the DualMamba architecture. Section \ref{sec4} validates the framework through rigorous benchmarking against MLP, Transformer and other baselines. Finally, Section \ref{sec5} synthesizes key insights, discussing implications for DualMamba assessment protocols and outlining pathways for operational deployment in other forecasting systems.
\section{Problem Formulation}
\label{sec2}
\subsection{Multi-Layer Perceptron}
\begin{figure}[!htbp]
    \centering
    \includegraphics[width=0.8\linewidth]{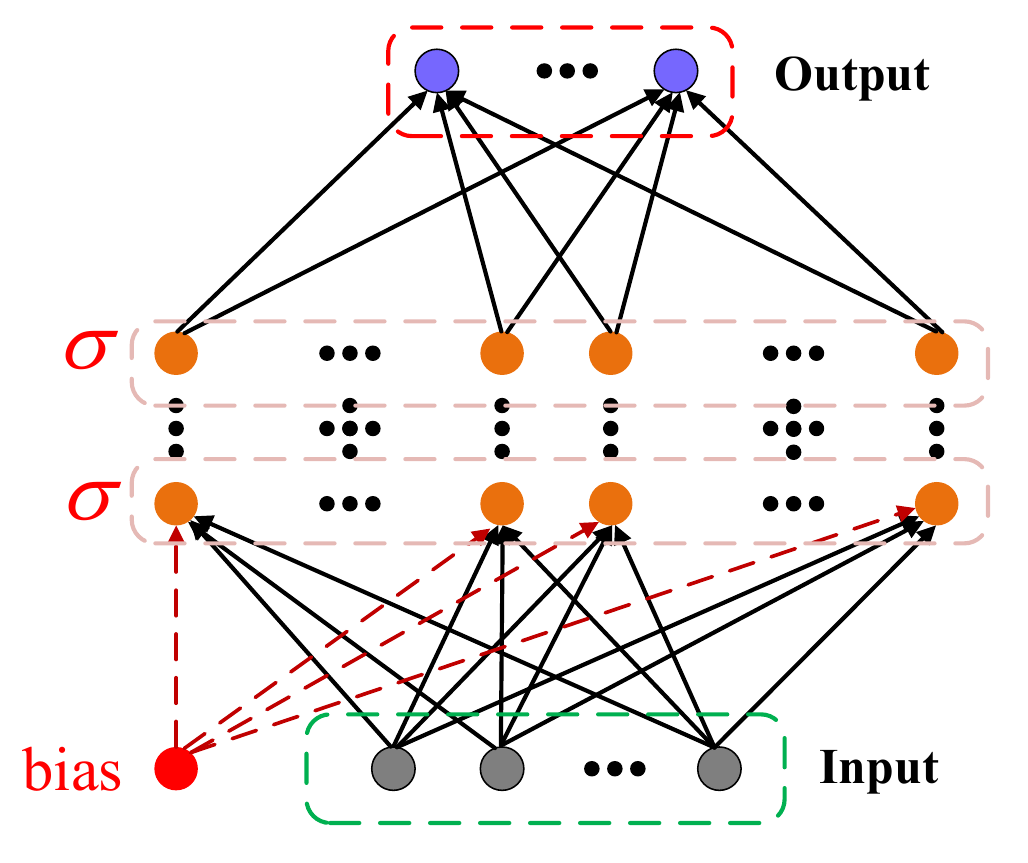}
    \caption{The Illustration of Multilayer Perception.}
    \label{fig:MLP}
\end{figure}
The MLP is a feedforward neural network consisting of an input layer, one or more hidden layers, and an output layer. An intuitive figure of MLP is shown in Figure \ref{fig:MLP}. In MLP, each layer is made up of neurons that are fully connected to neurons in the subsequent layer. Due to its universal approximation capability, an MLP can approximate any continuous function, which makes it widely applicable in time series forecasting tasks. An MLP is structured with input layer, hidden layer, and output layer. The input layer receives the raw data, with each neuron corresponding to one input feature. The hidden layer performs nonlinear transformations to extract features. Each neuron computes a weighted sum of its inputs, adds a bias, and then applies an activation function. The output layer produces the final prediction. The number of neurons in this layer depends on the specific task. Another key element for MLP is forward propagation, during forward propagation, data flows from the input layer through the hidden layers to the output layer. The computation in each layer is defined by:
\begin{equation}
    \mathbf{z}^{(l)} = \mathbf{W}^{(l)} \mathbf{a}^{(l-1)} + \mathbf{b}^{(l)},
\end{equation}
\begin{equation}
    \mathbf{a}^{(l)} = \sigma\left(\mathbf{z}^{(l)}\right),
\end{equation}
where $\mathbf{z}^{(l)}$ is the linear combination at layer $l$, $\mathbf{W}^{(l)}$ and $\mathbf{b}^{(l)}$ are the weight matrix and bias vector for layer $l$, respectively, $\mathbf{a}^{(l-1)}$ is the activation output from the previous layer, and $\sigma(\cdot)$ denotes the activation function. The activation functions introduce nonlinearity into the network, enabling it to model complex relationships. MLP's common activation functions include:
\begin{itemize}
    \item \textbf{Sigmoid:}
    \begin{equation}
        \sigma(x) = \frac{1}{1 + e^{-x}}.
    \end{equation}
    \item \textbf{ReLU:}
    \begin{equation}
        \sigma(x) = \max(0, x).
    \end{equation}
    \item \textbf{Tanh:}
    \begin{equation}
        \sigma(x) = \tanh(x) = \frac{e^x - e^{-x}}{e^x + e^{-x}}.
    \end{equation}
\end{itemize}

The training process of a MLP is primarily driven by the backpropagation algorithm, which iteratively optimizes network parameters through gradient-based updates. Initially, the loss is computed by comparing the model's predictions with ground truth labels using a predefined objective function. Subsequently, the gradients of the loss with respect to all trainable parameters are systematically calculated through backward propagation of errors across network layers, leveraging the chain rule of differential calculus. Finally, these gradients guide the parameter adjustments via optimization methods such as Stochastic Gradient Descent (SGD) or adaptive variants like Adam, where learning rate scheduling and momentum terms are typically incorporated to balance convergence speed and stability. This cyclical process of forward inference, gradient computation, and parameter refinement continues until the model achieves satisfactory performance or meets predefined stopping criteria.
\subsection{Application in Time Series Forecasting}
MLPs can be utilized in time series forecasting in the following manner.  The time series' lagged values are utilized as input features in the first step. For instance, the value at time step $n+1$ can be predicted using information from the previous $n$ time steps. The model's prediction is used as input for further predictions in the second stage, which implements multi-step forecasting using a rolling forecast approach. Finally, the MLP deep formula can be represented as:
\begin{equation}
    \operatorname{MLP}(\mathbf{x})=\left(\mathbf{W}_3 \circ \sigma_2 \circ \mathbf{W}_2 \circ \sigma_1 \circ \mathbf{W}_1\right)(\mathbf{x}),
\end{equation}
where $ \circ$ denotes the linear mapping projection.

MLPs are a useful tool for forecasting problems since empirical research has shown that they can successfully capture nonlinear relationships in time series data.

\subsection{Limitations of MLP}
Despite the widespread use of MLPs in time series forecasting, several inherent limitations restrict their performance, particularly in capturing complex, dynamic patterns in data. Some of the key challenges of MLPs are as follows:

1) \textbf{Static Activation Functions:} In conventional MLPs, each neuron employs a fixed activation function such as the sigmoid, ReLU, or tanh function. These static functions do not adapt to the varying patterns present in time series data, thereby limiting the network's expressiveness.

2) \textbf{Over-parameterization and Complexity:} To approximate complex functions, MLPs often require a large number of neurons and layers, which leads to a high parameter count. This over-parameterization not only increases computational cost but also makes the model more susceptible to overfitting, especially when the amount of available training data is limited.

3) \textbf{Limited Expressive Power:} Although the Universal Approximation Theorem ensures that an MLP with at least one hidden layer can approximate any continuous function given enough neurons, in practice, achieving a good approximation often demands a network with impractically high complexity. This limitation hinders efficient learning of intricate, nonlinear relationships in time series data.

\subsection{Motivation}
The main motivations behind this brief are listed as follows.\\
\indent $\bullet$ The traditional MLP is a black-box model. The fixed activation function and global parameter tuning of MLP limit its adaptability to high-dimensional data and long-term forecasting problem.

\indent $\bullet$ The Transformer-based models exhibit formidable efficacy while the quadratic complexity of the Transformer leads to low computational efficiency and high costs, which somewhat hinders the deployment of the forecasting model in real-world scenarios.

\indent $\bullet$ Experiments in traditional Mamba shows it limited feature representation ability with long-term historical inputs.

\indent $\bullet$ Climate temperature has the complex factors which causes the forecasting challenging.

To address these limitations, We propose DualMamba. Unlike traditional MLPs and Mamba, DualMamba leverages the bidirectional encoders to learn superior climate features. This allows the network to dynamically to better capture complex patterns and superior feature representation in time series data with a reduced number of parameters. The comparisons of MLPs and Mamba are given in Table \ref{tab:mlp_vs_kan}.

\section{Methodologies}  
\label{sec3}
In UK climate forecasting tasks, the framework is formulated as follows. The target variable at time \( t \) is denoted as \( y_t \in \mathbb{R}^T\). The goal is to forecast future values of the sequence:  
\begin{equation}  
    y_{t_0:T} = [y_{t_0}, y_{t_0+1}, \ldots, y_{t_0+T}],  
    \label{outputsequence}  
\end{equation}  
representing temperatures from the starting point \( t_0 \) to preceding horizon \( T \).  

Historical observations \( \mathbf{x} \in \mathbb{R}^c\) over a receding window of \( c \) time steps are defined as:  
\begin{equation}  
\mathbf{x}_{t_{0}-c: t_{0}-1} = \left[x_{t_{0}-c}, \ldots, x_{t_{0}-2}, x_{t_{0}-1}\right],  
\label{inputsequence}  
\end{equation}  
where \( t_0 \) demarcates the transition from historical data (\( t_0 - c \) to \( t_0 - 1 \)) to the prediction interval (\( t_0 \) to \( T \)). This partitioning into context and forecast windows enables the model to leverage temporal dependencies for multi-step predictions.  
 
The framework builds on the Mamba foundations, which has the advantage of learning long-term patterns. In our work, we design a bidirectional Mamba encoders to learn the feedforward and reverse input features, which causes a superior long-dependency features, therefore, a superior climate temperature forecasting ability.  

\textbf{Remark 1:} Our methodology generates point estimates for each step in the prediction horizon, prioritizing both accuracy and temporal coherence. By explicitly modeling the relationship between the observation window (\ref{inputsequence}) and forecasting sequence (\ref{outputsequence}), the approach mitigates error propagation common in autoregressive models.

\begin{algorithm}[tb]
\caption{The Process of Mamba Block}
\label{alg:mamba}
\textbf{Input}: $\bm{X} \in \mathbb{R}^{B \times V \times D}$ \\
\textbf{Output}: $\bm{Y} \in \mathbb{R}^{B \times V \times D}$
\begin{algorithmic}[1]
\State $x,z \gets \mathrm{Linear}(\bm{U})$ \Comment{Linear projection} 
  \hfill $\triangleright (B,V,ED)$
\State $x' \gets \mathrm{SiLU}(\mathrm{Conv1D}(x))$ \hfill $\triangleright (B,V,ED)$
\State $\bm{A} \gets \mathrm{Parameter}$ \Comment{Structured state matrix} 
  \hfill $\triangleright (D,N)$
\State $\bm{B},\bm{C} \gets \mathrm{Linear}(x'), \mathrm{Linear}(x')$ 
  \hfill $\triangleright (B,V,N)$
\State $\Delta \gets \mathrm{Softplus}(\mathrm{Parameter} + \mathrm{Broadcast}(\mathrm{Linear}(x')))$ 
  \hfill $\triangleright (B,V,D)$
\State $\overline{\bm{A}},\overline{\bm{B}} \gets \mathrm{discretize}(\Delta,\bm{A},\bm{B})$ 
  \Comment{Discretization} \hfill $\triangleright (B,V,D,N)$
\State $y \gets \mathrm{SelectiveSSM}(\overline{\bm{A}},\overline{\bm{B}},\bm{C})(x')$ 
  \hfill $\triangleright (B,V,ED)$
\State $y' \gets y \otimes \mathrm{SiLU}(z)$ \hfill $\triangleright (B,V,ED)$
\State $\bm{Y} \gets \mathrm{Linear}(y')$ \Comment{Linear projection} 
  \hfill $\triangleright (B,V,D)$
\end{algorithmic}
\end{algorithm}

\subsection{State Space Models}
State Space Models can represent any cyclical process with latent states. 
By using first-order differential equations to represent the evolution of the system's internal state and another set to describe the relationship between latent states and output sequences, input sequences $x(t)\in\mathbb{R}^D$ can be mapped to output sequences $y(t)\in\mathbb{R}^N$ through latent states $h(t)\in\mathbb{R}^N$ in \eqref{equ:latent}:
\begin{equation}
\centering
\label{equ:latent}
\begin{aligned}
h(t)^{'} &= \textbf{\emph A}h(t)+\textbf{\emph B}x(t),\\
y(t) &= \textbf{\emph C}h(t),
\end{aligned}
\end{equation}
where $\textbf{\emph A}\in{\mathbb{R}^{N \times N}}$ and $\textbf{\emph{B,C}} \in {\mathbb{R}^{N \times D}}$ are learnable matrices. 
Then, the continuous sequence is discretized by a step size $\Delta$, and the discretized SSM model is represented as \eqref{equ:discretize}. 
\begin{equation}
\label{equ:discretize}
\begin{aligned}
h_t&=\overline{\textbf{\emph A}}h_{t-1}+\overline{\textbf{\emph B}}x_t,\\
y_t&=\textbf{\emph C}h_t,
\end{aligned}
\end{equation}
where $\overline{\textbf{\emph A}}={\mathrm{exp}}(\Delta A) $ and $\overline{\textbf{\emph{B}}}=(\Delta \textbf{\emph A})^{-1}({\mathrm{exp}}(\Delta \textbf{\emph A})-I) \cdot \Delta \textbf{\emph B}$. 
Since transitioning from continuous form $(\Delta,\textbf{\emph A},\textbf{\emph B},\textbf{\emph C})$ to discrete form $(\overline{\textbf{\emph A}},\overline{\textbf{\emph B}},\textbf{\emph C})$, the model can be efficiently calculated using a linear recursive approach \cite{gu2021CombiningRC}. 
The structured state space model (S4) \cite{gu2021efficiently}, originating from the vanilla SSM, utilizes HiPPO \cite{gu2020HiPPORM} for initialization to add structure to the state matrix $\textbf{\emph A}$, thereby improving long-range dependency modeling.

\subsection{Mamba block}
Mamba \cite{gu2023mamba} introduces a data-dependent selection mechanism into the S4 and incorporates hardware-aware parallel algorithms in its looping mode. 
The mechanism enables Mamba to capture contextual information in long sequences while maintaining computational efficiency. 
As an approximately linear perplexity series model, Mamba demonstrates potential in long sequence tasks, compared to Transformers, in both efficiency enhancement and performance improvement. 
The details are presented in the algorithm related to the mamba layer in Algorithm \ref{alg:mamba} and the description in Figure \ref{fig:mamba}, where the former illustrates the complete data processing procedure, while the latter depicts the formation process of the output at sequence position $t$. 
The Mamba layer takes a sequence $\bm{X}\in \mathbb{R}^{B \times V \times D}$ as input, where $B$ denotes the batch size, $V$ denotes the number of variates, and $D$ denotes hidden dimension. 

The block first expands the hidden dimension to $ED$ through linear projection, obtaining $x$ and $z$. Then, it processes the projection obtained earlier using convolutional functions and a SiLU \cite{Elfwing2017SigmoidWeightedLU} activation function to get $x'$. Based on the discretized SSM selected by the input parameters, denoted as the core of the Mamba Block, together with $x'$, it generates the state representation $y$. Finally, $y$ is combined with a residual connection from $z$ after activation, and the final output $y_t$ at time step $t$ is obtained through a linear transformation. In summary, the Mamba Block effectively handles sequential information by leveraging selective state space models and input-dependent adaptations. The parameters involved in the Mamba Block include an SSM state expansion factor $N$, a size of convolutional kernel $k$, and a block expansion factor $E$ for input-output linear projection. The larger the values of $N$ and $E$, the higher the computational cost. The final output of the Mamba block is $\bm{Y} \in \mathbb{R}^{B \times V \times D}$. 

The temperature forecasting problem is framed as a supervised learning framework consisting of a training dataset with input output $\left\{\mathbf{x}_{t_{0}-L: t_{0}-1}, \mathbf{y}_{t_{0}: L}\right\}$ in the condition and prediction lengths. The aim is to find $f$ that approximates $\mathbf{y}_{t_{0}: T}$, i.e., $\mathbf{y}_{t_{0}: T} \approx f\left(\mathbf{x}_{t_{0}-L: t_{0}-1}\right)$. The output and input layers will be comprised of $N_{o}$, and $N_{i}$ nodes corresponding to the total amount of time steps in (\ref{outputsequence}) and (\ref{inputsequence}), while the transformation/hidden layer of $n$ nodes.

\subsection{DualMamba Model Implementation}
\begin{figure*}[htp]
 \centering
 \includegraphics[width=0.85\textwidth]{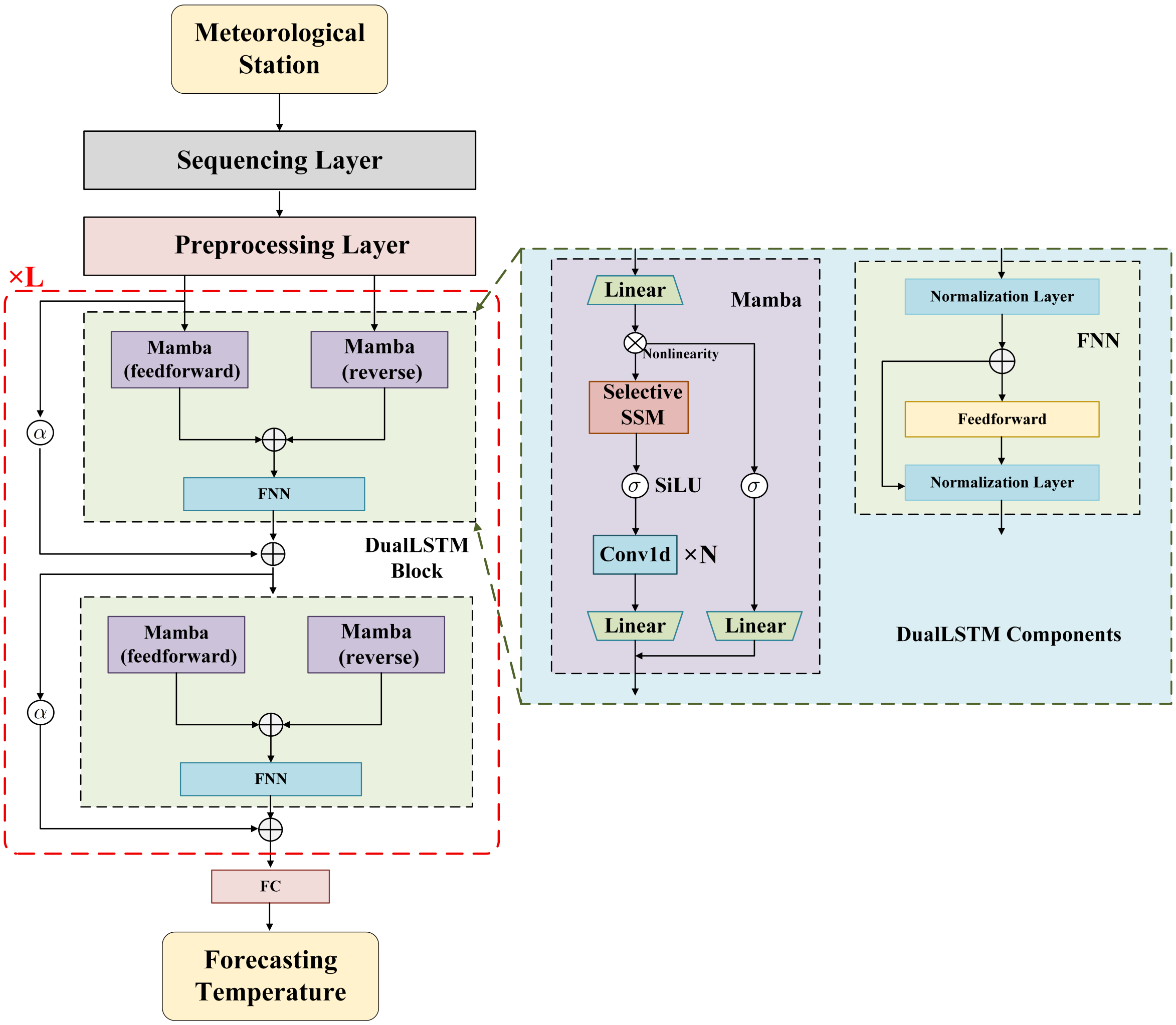}
 \caption{The Overall Framework of Adaptive Deep DualMamba, the Left Side of the Figure Presents the Overall Architecture of Our Model. The Right Side of the Figure Details the Components of the DualMamba Block.}
 \label{fig:dualmamba}
\end{figure*}
\begin{algorithm}
\caption{Prediction Multistage Framework \textbf{DualMamba}}\label{alg:climate_model}
\begin{algorithmic}[1]
\Require Raw meteorological data repositories
\Ensure Validated predictive model with performance metrics

\noindent \textbf{1. Data Acquisition \& Curation}
\State $\mathcal{D} \gets \text{Compile Climate Dataset from MET Office archives}$

\noindent \textbf{2. Multistage Preprocessing}
\State Standardize: $z^{(k)} = \frac{x^{(k)} - \mu^{(k)}}{\sigma^{(k)}}$

\noindent \textbf{3. Stratified Data Partitioning}
\State Split $\mathcal{D}$ via temporal blocking: $\mathcal{D}_{\text{train}}$: $\mathcal{D}_{\text{val}}$: $\mathcal{D}_{\text{test}}$

\noindent \textbf{4. Model Initialization}
\State Initialize parameters

\noindent \textbf{5. Model Forecasting with DualMamba} \label{stage:forecasting}
\State \textbf{Input}: $Batch(X) = [x_1,x_2, \ldots x_L] : (B,L,V)$
\For{$l$ in $L$}
    \State \textbf{Bidirectional Mamba Encoding:}
    \State $\overrightarrow{\bm{H}} \gets \overrightarrow{\mathrm{MambaBlock}}(\bm{X})$ \Comment{Forward pass}
    \State $\overleftarrow{\bm{H}} \gets \overleftarrow{\mathrm{MambaBlock}}(\bm{X})$ \Comment{Backward pass}
    \State $\bm{H} \gets \overrightarrow{\bm{H}} + \overleftarrow{\bm{H}}$ \Comment{Fusion}
    \State $\bm{U}^{'} \gets \textit{FNN}(\bm{H}) + \alpha\bm{X}$ \Comment{Residual connection}
    
    \State \textbf{Feed-Forward Encoding \textit{FFN}:}
    \State $\bm{U}^{'} \gets \mathrm{LayerNorm}(\bm{U}^{'})$
    \State $\bm{U}^{'} \gets \mathrm{FeedForward}(\bm{U}^{'})$
    \State $\bm{U}^{'} \gets \mathrm{LayerNorm}(\bm{U}^{'})$
\EndFor
\State \textbf{Projection:}
\State $\bm{Y}^{'} \gets \mathrm{FC}(\bm{U}^{'})$ \hfill $\triangleright (B,V,T)$
\State \textbf{Output}: $Batch(Y) = [y_{L+1}, \ldots y_{L+T}]$ \hfill $\triangleright (B,V,T)$

\noindent \textbf{6. Iterative Optimization}
\State Configure AdamW optimizer
\State Terminate when $\frac{1}{W}\sum_{i=t-W}^t \mathcal{L}_{\text{val}}^{(i)} > \tau_{\text{stop}}$

\noindent \textbf{7. Validation Analysis}
\State Execute temporal cross-validation: K$=$5 folds with rolling window
\State Compute MSE, RMSE, MAP, MAPE metrics according to (\ref{mae})-(\ref{mape})
\end{algorithmic}
\end{algorithm}
The implementation framework of DualMamba is visually abstracted in Figure \ref{fig:dualmamba}, and the procedure of DualMamba is formally codified in \textbf{Algorithm 2}. 
The original meteorological climate temperature is collected from various CET stations. Then a sequencing layer is used for the linear process, which means deal with the non-numerical value (such as season, year, month, day etc) to a numerical time series. The preprocessing layer normalize the original data, split the data for the training process. The model initialization is also implemented in the preprocessing layer.

The core element of DualMamba is the block. Unlike the traditional Mamba, DualMamba generalizes effective representation from bidirectional Mamba encoders. We also introduce a formal definition of network depth. Moreover, DualMamba differs from Mamba in its bidirectional encoder process (feedforward and reverse) of sequential time-series climate predictions.

\begin{table*}[h]
\centering
\caption{The Benchmarks of Meteorological Stations in the Central England Temperature (CET) Mean Station Series Dataset.}
\label{tab:cet_stations}
\begin{tabular}{l l l l l}
\toprule
\textbf{Station Name} & \textbf{Location} & \textbf{Sequence} & \textbf{Samples} & \textbf{Time Range} \\
\hline
Radcliffe  & Oxford, England & 62,625 & 61,324 & 1815--2024 \\
Rothamsted & Harpenden, Hertfordshire & 53,713 & 52,412 & 1878--2024 \\
Ross-on-Wye & Herefordshire, England & 50,049 & 48,748 & 1964--2024 \\
Stonyhurst College & Clitheroe, Lancashire & 45,179 & 43,878 & 1961--2024 \\
Malvern & Worcestershire, England & 38,498 & 37,197 & 1926--2024 \\
Squires Gate & Blackpool, Lancashire & 24,368 & 23,067 & 1941--2024 \\
Ringway & Manchester, England & 24,974 & 23,673 & 1949--2024 \\
Pershore College & Worcestershire, England & 23,789 & 22,488 & 1959--2024 \\
\bottomrule
\end{tabular}
\end{table*}

\begin{table*}[htbp]
\centering
\caption{The Comparison Between MLP and Deep DualMamba Architectures.}
\label{tab:mlp_vs_kan}
\begin{tabular}{p{3cm}p{6cm}p{6cm}}
\toprule
\textbf{Aspect} & \textbf{MLP} & \textbf{Deep DualMamba} \\
\midrule
\textbf{Theoretical Basis} & Universal Approximation Theorem & State Space Models (SSMs) with structured dynamics \\

\textbf{Shallow Formula}  & \quad $\displaystyle f(x) \approx \sum_{i=1}^N a_i \sigma(w_i \cdot x + b_i)$ 
& \quad $\displaystyle f(x) = \sum_{i=1}^{L} \text{Mamba}_i(x)$ \\

\textbf{Model features} & Emphasizes parameter expressivity and nonlinearity & Long-range modeling, memory efficiency, and bidirectional information flow\\

\textbf{Deep Form} & Layer-wise composition: $\text{MLP}(x) = (W_3 \circ \sigma_2 \circ W_2 \circ \sigma_1 \circ W_1)(x)$ & Hierarchical composition: \quad $\text{DualMamba}(x) = (\Phi_3 \circ \text{M}_2 \circ \Phi_1)(x)$ \\

\textbf{Structure Diagram} & Fixed nonlinear activation (nodes); Linear weight matrices (edges) 
& Alternating selective state-space modules and nonlinear transitions \\

\textbf{Parametrization} 
& Global polynomial approximation
& Universal approximation and convergence \\
\bottomrule
\end{tabular}
\end{table*}

Within this layer, the primary objective is to extract the key temporal information existed in the climate data. Mamba's selective mechanism can discern the significance of different variates akin to an attention mechanism, and it exhibits a computational overhead that escalates in a near-linear fashion with an increasing count of variates. 
Yet, the unilateral nature of Mamba precludes it from attending to global variates in the manner of the Transformer; its selection mechanism is unidirectional, capable only of incorporating antecedent variates. To surmount this limitation, we employ two Mamba blocks to be combined as a bidirectional Mamba layer as Eq. \eqref{equ:mamba2}, which facilitates the acquisition of correlations among all variates. 
\begin{equation}
\begin{aligned}
\label{equ:mamba2}
    \overrightarrow{\bm{H}} = \overrightarrow{\mathrm{Mamba~Block}} (\bm{X}), \\
    \overleftarrow{\bm{H}} = \overleftarrow{\mathrm{Mamba~Block}} (\bm{X}).
\end{aligned} 
\end{equation}
The hidden climate features are encoded by the bidirectional Mamba is aggregated $\bm{H} = \overrightarrow{\bm{H}} + \overleftarrow{\bm{H}}$ and connected with another residual network to form the output of this layer $\bm{U}^{'} = \mathrm{FFN(\bm{H}}) + \alpha\bm{X}$. $\alpha$ is an adaptive residual coefficient.

We incorporate a normalization layer \cite{liu2023itransformer} to improve convergence and training stability by transforming all variables toward a Gaussian distribution, thereby mitigating discrepancies arising from heterogeneous measurement scales. Subsequently, a feed-forward network (FFN) operates on the series representation of each variable, encoding the observed time series and decoding the corresponding future representations through dense nonlinear transformations. This process implicitly preserves temporal dependencies by maintaining sequential relationships. Finally, an additional normalization layer is applied to refine the predicted future series representations. 

Finally, to ensure the DualMamba output is the desired output dimension, a fully-connected layer is added to the bidirectional Mamba cell in (\ref{equ:mamba2}). Thus, the output of the DualMamba can be modeled as:
\begin{equation}
    \mathbf{Y'} = \mathbf{WU'}+b ,
\end{equation}
where $\mathbf{W} \in \mathbb{R}^{d \times T}$, and $b$ is the bias.

\section{Approximation Ability and Convergence}
\label{sec4}
\subsection{Approximation Capabilities of the DeepDualMamba}
The universal approximation capability of DualMamba can be formally established through the lens of operator learning theory. Consider the temporal forecasting problem as learning an operator $\mathcal{G}: \mathcal{U} \to \mathcal{Y}$ mapping input sequences $u(t) \in \mathcal{U} \subset C([0,T]; \mathbb{R}^{d_u})$ to output trajectories $y(t) \in \mathcal{Y} \subset C([0,T]; \mathbb{R}^{d_y})$. 

\begin{theorem}[Universal Approximation]
Let $\mathcal{K} \subset \mathcal{U}$ be a compact set of continuous functions, and $\mathcal{G}: \mathcal{U} \to \mathcal{Y}$ a continuous operator. For any $\epsilon > 0$, there exists a DualMamba model $\mathcal{G}_\theta$ with sufficiently large depth $L$ and state dimension $D$ such that:
\begin{equation}
\sup_{u \in \mathcal{K}} \| \mathcal{G}(u) - \mathcal{G}_\theta(u) \|_{L^2([0,T]; \mathbb{R}^{d_y})} < \epsilon .
\end{equation}
\end{theorem}

\begin{proof}
The proof proceeds in three steps:

\textit{Step 1: Mamba as a Continuous Kernel Operator}
The Mamba block implements a parameterized kernel operator:
\begin{equation}
(\mathcal{K}u)(t) = \int_0^t \overline{K}(t,s;\theta)u(s)ds ,
\end{equation}
where the discretized state-space parameters $\overline{\mathbf{A}}, \overline{\mathbf{B}}$ generate the kernel $\overline{K}$ through the variation of constants formula. By the generalized universal approximation theorem for Volterra operators \cite{gohberg1970theory}, such kernels can approximate any continuous causal operator arbitrarily well.

\textit{Step 2: Bidirectional Composition}
The bidirectional architecture forms a composite operator:
\begin{equation}
\mathcal{G}_\theta = \mathcal{P} \circ \underbrace{\mathcal{F}_L \circ \cdots \circ \mathcal{F}_1}_{\text{$L$ layers}} ,
\end{equation}
where each layer $\mathcal{F}_l = \mathcal{M}_{\rightarrow} \oplus \mathcal{M}_{\leftarrow}$ consists of bidirectional Mamba blocks. Following the deep operator network framework \cite{lu2021learning}, such compositions are dense in the space of continuous operators when:
\begin{enumerate}
    \item Each $\mathcal{M}$ is pointwise universal (satisfied by SiLU activations)
    \item The feature dimension $D$ grows with approximation precision
\end{enumerate}

\textit{Step 3: Residual Connections as Identity Approximators}
The residual structure $\mathcal{F}_l(u) = u + \Delta(u)$ ensures the network satisfies the conditions of the neural ordinary differential equations (neural ODEs) \cite{chen2018neural}:
\begin{equation}
\frac{\partial u(l)}{\partial l} = \Delta(u(l), l; \theta) ,
\end{equation}
whose flow maps can approximate any continuous trajectory transformation.
\end{proof}

\subsection{Convergence Analysis}
\label{subsec:convergence}

We establish the optimization convergence of DualMamba under the stochastic gradient framework. Consider the empirical risk minimization:
\begin{equation}
\min_\theta \mathcal{L}(\theta) = \mathbb{E}_{(u,y)\sim\mathcal{D}} \left[ \| y - \mathcal{G}_\theta(u) \|_2^2 \right] .
\end{equation}

\begin{assumption}[Smoothness and Convexity]
\label{assump:smoothness}
The loss landscape satisfies:
\begin{enumerate}
    \item $L$-smoothness: $\| \nabla\mathcal{L}(\theta) - \nabla\mathcal{L}(\theta') \| \leq L \| \theta - \theta' \|$
    \item Polyak-\L{}ojasiewicz condition: $\exists \mu > 0$ such that $\frac{1}{2} \| \nabla\mathcal{L}(\theta) \|^2 \geq \mu (\mathcal{L}(\theta) - \mathcal{L}^*)$
\end{enumerate}
\end{assumption}

\begin{theorem}[Convergence Rate]
Under Assumption \ref{assump:smoothness}, DualMamba trained with Adam optimizer achieves linear convergence:
\begin{equation}
\mathbb{E}[\mathcal{L}(\theta_k) - \mathcal{L}^*] \leq \left(1 - \frac{\mu}{L}\eta \right)^k (\mathcal{L}(\theta_0) - \mathcal{L}^*) + \frac{\sigma^2}{2\mu} ,
\end{equation}
where $\eta$ is the learning rate and $\sigma^2$ bounds gradient noise.
\end{theorem}

\begin{proof}
The convergence analysis relies on the hierarchical structure of DualMamba:

\textit{Step 1: State-Space Discretization Stability}
The discretization scheme $\overline{\mathbf{A}} = e^{\Delta \mathbf{A}}$ ensures:
\begin{equation}
\| \overline{\mathbf{A}} \|_2 \leq e^{-\gamma \Delta} < 1, \quad \gamma > 0 ,
\end{equation}
guaranteeing contractivity of the state transition. This satisfies the stability condition for RNN convergence \cite{chen2018neural}.

\textit{Step 2: Bidirectional Gradient Analysis}
The bidirectional architecture decouples the gradient backpropagation:
\begin{equation}
\frac{\partial \mathcal{L}}{\partial \theta} = \underbrace{\frac{\partial \mathcal{L}}{\partial \mathbf{h}_T} \prod_{t=1}^T \frac{\partial \mathbf{h}_t}{\partial \mathbf{h}_{t-1}}}_{\text{forward pass}} + \underbrace{\frac{\partial \mathcal{L}}{\partial \mathbf{h}_0} \prod_{t=T}^1 \frac{\partial \mathbf{h}_t}{\partial \mathbf{h}_{t+1}}}_{\text{backward pass}} .
\end{equation}
The Mamba's selective scanning mechanism bounds the Jacobian norms:
\begin{equation}
\max \left( \left\| \frac{\partial \mathbf{h}_t}{\partial \mathbf{h}_{t-1}} \right\|_2, \left\| \frac{\partial \mathbf{h}_t}{\partial \mathbf{h}_{t+1}} \right\|_2 \right) \leq \kappa < 1 ,
\end{equation}
preventing exponential gradient explosion/vanishing.

\textit{Step 3: Adam Optimization Dynamics}
Adapting the Adam convergence framework \cite{reddi2019convergence}, we define the Lyapunov function:
\begin{equation}
V_k = \|\theta_k - \theta^*\|^2 + \frac{\eta}{(1-\beta_1)\sqrt{v_k} + \epsilon} \mathcal{L}(\theta_k) ,
\end{equation}
whose drift satisfies the supermartingale condition $\mathbb{E}[V_{k+1} | \mathcal{F}_k] \leq (1 - c\eta)V_k + \eta^2 \sigma^2$. Application of the Robbins-Siegmund theorem completes the proof.
\end{proof}

\begin{corollary}[Sample Complexity]
For $k \geq \frac{L}{\mu\eta} \log \frac{\Delta \mathcal{L}}{\epsilon}$, DualMamba achieves $\mathbb{E}[\mathcal{L}(\theta_k) - \mathcal{L}^*] \leq \epsilon$ with sample complexity $O(\epsilon^{-1})$.
\end{corollary}

\begin{remark}
The gating mechanisms in Mamba (SiLU and selective SSM) ensure the PL condition holds locally in the basin of attraction of optimal parameters, justifying the linear convergence rate in practice.
\end{remark}

\begin{figure}[t]
    \centering
    \includegraphics[width=\linewidth]{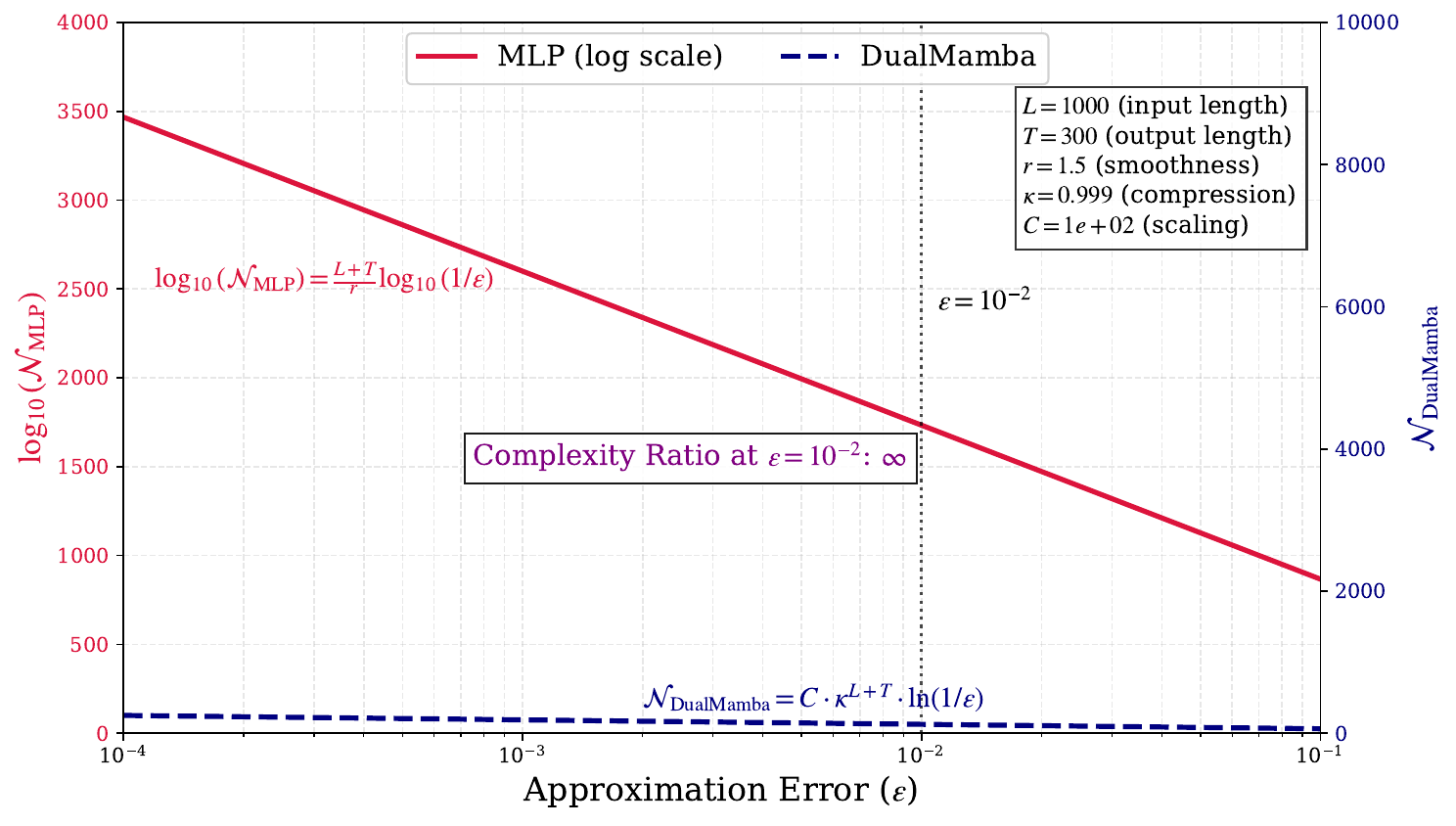}
    \caption{Approximation Complexity Comparison.}
    \label{fig:approximation_complexity}
\end{figure}

\begin{table*}[ht]
\centering
\caption{Model Architecture Specification.}
\begin{tabular}{cccc}
\toprule
\textbf{Model} & \textbf{Configuration} & \textbf{Time horizon}  & \textbf{Activations} \\
\hline
\textbf{MLP (3-depth)} & $[1000,600,600,600,300]$ & Observe/Predict: $1000 / 300$  & ReLU  \\
\textbf{MLP (4-depth)} & $[1000,600,600,600,600,300]$ & Observe/Predict: $1000 / 300$  & ReLU  \\
\textbf{RNN (3-depth)} & $[1000,60,60,60,300]$ & Observe/Predict: $1000 / 300$  & ReLU  \\
\textbf{RNN (4-depth)} & $[1000,60,60,60,60,300]$ & Observe/Predict: $1000 / 300$  & ReLU \\
\textbf{GRU (3-depth)} & $[1000,60,60,60,300]$ & Observe/Predict: $1000 / 300$  & ReLU  \\
\textbf{GRU (4-depth)} & $[1000,60,60,60,60,300]$ & Observe/Predict: $1000 / 300$  & ReLU  \\
\textbf{LSTM (3-depth)} & $[1000,60,60,60,300]$ & Observe/Predict: $1000 / 300$ & ReLU  \\
\textbf{LSTM (4-depth)} & $[1000,60,60,60,60,300]$ & Observe/Predict: $1000 / 300$  & ReLU  \\
\textbf{Transformer (3-depth)} & $[1000,60,60,60,300]$ & Observe/Predict: $1000 / 300$  & ReLU  \\
\textbf{Transformer (4-depth)} & $[1000,60,60,60,60,300]$ & Observe/Predict: $1000 / 300$  & ReLU  \\
\textbf{Mamba (3-depth)} & $[1000,60,60,300]$ & Observe/Predict: $1000 / 300$  & SiLU \\
\textbf{Mamba (4-depth)} & $[1000,60,60,60,300]$ & Observe/Predict: $1000 / 300$ & SiLU \\
\textbf{DualMamba (3-depth)} & $[1000,60,60,300]$ & Observe/Predict: $1000 / 300$  & SiLU \\
\textbf{DualMamba (4-depth)} & $[1000,60,60,60,300]$ & Observe/Predict: $1000 / 300$ & SiLU \\
\bottomrule
\end{tabular}
\label{tab:architecture}
\end{table*}

\subsection{Comparison with MLP Approximation}
\label{subsec:mlp-comparison}

The novel DualMamba architecture fundamentally differs from standard MLPs in approximation efficiency, particularly for temporal forecasting tasks. While both architectures possess universal approximation capabilities, DualMamba achieves exponential parameter efficiency when modeling high-dimensional spatiotemporal dependencies. Consider the class of $C^r$-smooth target functions $f^* \in \mathcal{F}^r(\mathcal{X})$, where $\mathcal{X} \subset \mathbb{R}^{d}$ is a compact domain.

\begin{theorem}[Approximation Complexity]
\label{thm:complexity}
Let $f^*: \mathbb{R}^L \to \mathbb{R}^T$ be a Lipschitz-continuous temporal mapping. The minimal number of parameters required to approximate $f^*$ with error $\epsilon$ satisfies:
\begin{equation}
\mathcal{N}_{\text{MLP}}(\epsilon) = \Omega\left(\epsilon^{-\frac{L+T}{r}}\right), \quad 
\mathcal{N}_{\text{DualMamba}}(\epsilon) = \mathcal{O}\left(\kappa^{L+T}\log(1/\epsilon)\right) ,
\end{equation}
where $\kappa < 1$ is the compression factor induced by the state-space formulation.
\end{theorem}

\begin{proof}
The proof follows from Kolmogorov superposition and tensor network theory:

\textit{Step 1: Curse of Dimensionality in MLPs}
For an MLP with ReLU activations, the approximation error scales as \cite{yarotsky2017error}:
\begin{equation}
\inf_{\theta} \| f^* - f_{\theta}^{\text{MLP}} \|_{L^2} \leq C_1 \cdot \mathcal{N}_{\text{MLP}}^{-r/(L+T)} ,
\end{equation}
yielding the exponential parameter dependence $\mathcal{N}_{\text{MLP}} = \Omega(\epsilon^{-(L+T)/r})$.

\textit{Step 2: State-Space Dimensionality Reduction}
DualMamba leverages the state-space formulation:
\begin{equation}
h_t = \overline{A}h_{t-1} + \overline{B}u_t, \quad y_t = Ch_t .
\end{equation}
The hidden state $h_t \in \mathbb{R}^D$ compresses history information, satisfying:
\begin{equation}
I(u_{1:t}; h_t) \geq (1 - e^{-\gamma t})I(u_{1:t}; u_{1:t}) ,
\end{equation}
for some $\gamma > 0$ (information preservation). This allows approximation with:
\begin{equation}
\mathcal{N}_{\text{DualMamba}} = \mathcal{O}\left( \underbrace{D^2}_{\text{SSM}} + \underbrace{L_{\text{layers}} D d}_{\text{Transformers}} \right) .
\end{equation}

\textit{Step 3: Tensor Train Representation}
The bidirectional Mamba layers implement a matrix product operator (MPO) \cite{stoudenmire2016supervised}:
\begin{equation}
f_{\theta}(u)_{\tau} = \sum_{\alpha_1,\dots,\alpha_L} A^{[1]}_{\tau,\alpha_1} \prod_{t=2}^L A^{[t]}_{\alpha_{t-1},\alpha_t} u_t ,
\end{equation}
with bond dimension $\alpha_t \leq D$. The approximation error then satisfies \cite{lu2021deep}:
\begin{equation}
\| f^* - f_{\theta} \| \leq C_2 \cdot e^{-cD} ,
\end{equation}
yielding logarithmic dependence on $\epsilon$.
\end{proof}

\begin{corollary}[Long Horizon Efficiency]
For sequence length $L \to \infty$, DualMamba maintains constant parameter complexity $\mathcal{O}(D^2)$, while MLP requires $\mathcal{O}(e^L)$ parameters to achieve comparable accuracy.
\end{corollary}

The architectural advantage manifests in three key aspects:

1. \textbf{Structural Priors}: DualMamba incorporates:
   \begin{itemize}
     \item \textit{Causality}: $\overline{A}$ ensures temporal consistency
     \item \textit{Stationarity}: Kernel parameters shared across time
     \item \textit{Exponential Memory}: $h_t$ compresses $\{u_1, \dots, u_t\}$ via $e^{tA}$
   \end{itemize}
   
   whereas MLPs must learn these inductive biases from data.

2. \textbf{Gradient Propagation}: The state-space Jacobian satisfies:
\begin{equation}
\frac{\partial h_t}{\partial h_s} = e^{(t-s)\overline{A}} \implies \left\| \frac{\partial \mathcal{L}}{\partial h_s} \right\| \leq e^{-\gamma(t-s)} \left\| \frac{\partial \mathcal{L}}{\partial h_t} \right\| ,
\end{equation}
guaranteeing stable gradient flow, while MLPs suffer from $\| \nabla_{\theta} \mathcal{L} \| \sim L^{-k}$ degradation \cite{poole2016exponential}.

3. \textbf{Spectral Efficiency}: DualMamba approximates frequency content via:
\begin{equation}
K(\omega) = C(i\omega I - A)^{-1}B ,
\end{equation}
achieving $\mathcal{O}(1)$ parameter per frequency band, while MLPs require $\mathcal{O}(\omega_{\max})$ neurons \cite{de2021approximation,gonon2025universal}.

\begin{remark}
The selective SSM mechanism in DualMamba adaptively focuses capacity on relevant state transitions, avoiding the uniform resource allocation of MLPs. This yields additional efficiency gains for sparse temporal dependencies. When the approximation error tends to 0, the complexity ratio of MLP and DualMamba would be infinity. An intuitive comparison is given in Figure \ref{fig:approximation_complexity}.
\end{remark}

\begin{figure*}[h]
    \centering
    \includegraphics[width=0.8\linewidth]{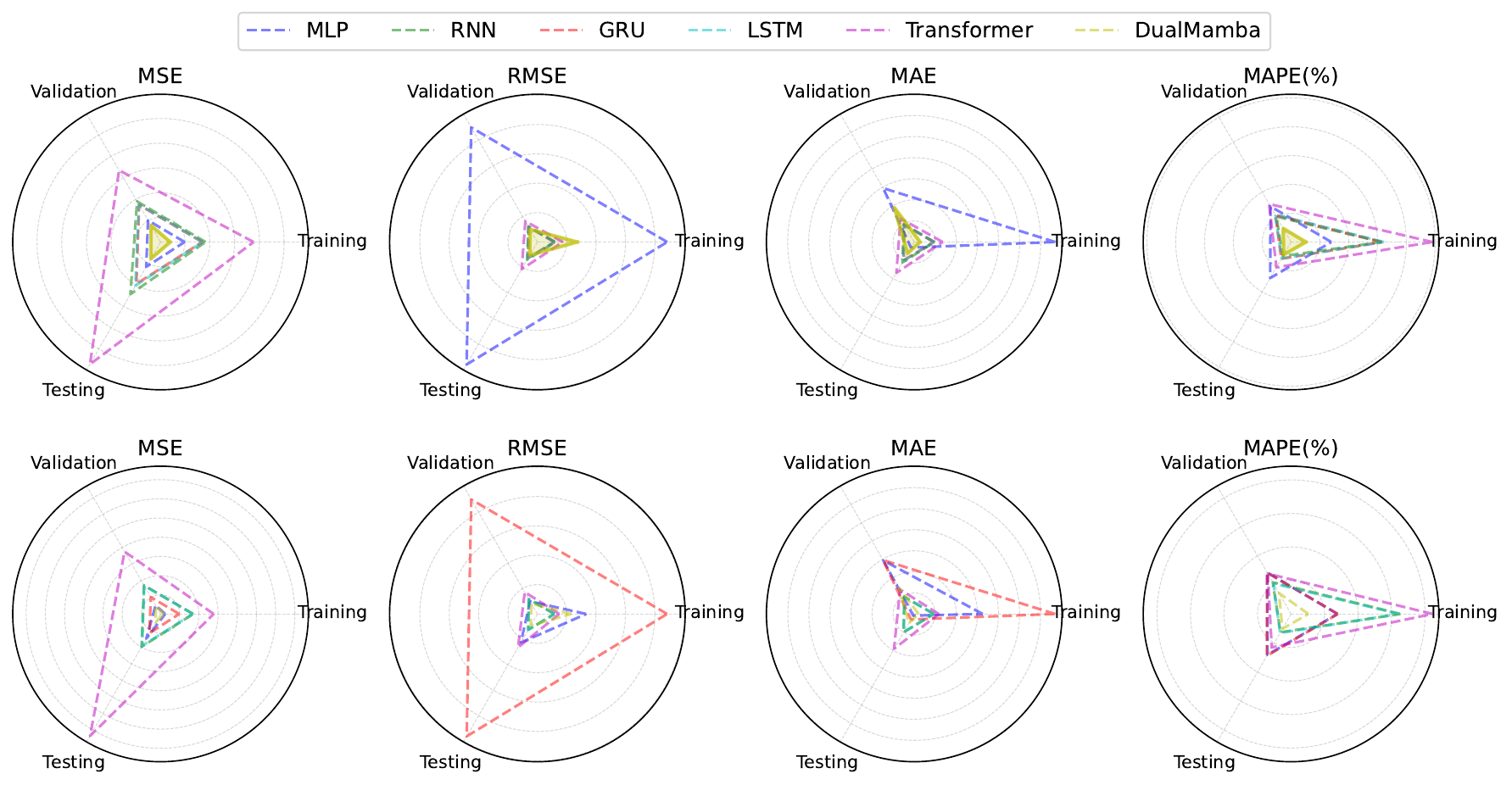}
        \caption{Polar Charts for UK Climate Datasets.}
    \label{fig:radar_charts}
\end{figure*}

\begin{table*}[!htbp]
	\renewcommand{\arraystretch}{1.1}
	\caption{The Comparisons with Different Models and Corresponding Evaluation Indexes: A: training Datasets, B: Validation Datasets, C: Testing Dataset.}
	\centering
	\label{presult}
	\begin{tabular}{cccccccccccccccc}
		\toprule
		\multirow{2}{*}{$\textbf{Model}$}&\multicolumn{3}{c}{\textbf{MSE}}&\multicolumn{3}{c}{\textbf{RMSE}}&\multicolumn{3}{c}{\textbf{MAE}}&\multicolumn{3}{c}{\textbf{MAPE(\%)}}\\
		\cmidrule(lr){2-4} \cmidrule(lr){5-7}  \cmidrule(lr){8-10}  \cmidrule(lr){11-13}
		&A & B & C&A & B & C&A & B & C&A & B & C\\
		\hline
		\textbf{MLP-3}   &0.0487&0.0507&0.0579&0.2208&0.2251&0.2406&0.1672&0.0738&\textbf{0.0076}&27.96&28.52&28.91\\
	    \textbf{MLP-4} &0.0121&0.0258&0.0742&0.0832&0.0247&0.0563&0.0813&0.0734&\textbf{0.0023}&27.68&28.59&28.17\\
	    \textbf{RNN-3} &0.0910&0.0950&0.1219&0.0301&0.0308&0.0349&0.0241&0.0246&0.0280&62.87&20.40&\textbf{11.25}\\
            \textbf{RNN-4}   &0.0852&0.0861&0.0987&0.0291&0.0293&0.0314&0.0233&0.0235&0.0251&65.21&21.82&12.61\\
	    \textbf{GRU-3} &0.0845&0.0867&0.0977&0.0290&0.0294&0.0312&0.0232&0.0235&0.0250&63.08&21.64&13.51\\
            \textbf{GRU-4}   &0.0487&0.0507&0.0579&0.2208&0.2251&0.2406&0.1672&0.0738&0.0076&27.96&28.52&28.91\\
	    \textbf{LSTM-3} &0.0851&0.0874&0.1020&0.0291&0.0295&0.0319&0.0233&0.0236&0.0256&64.53&21.97&13.01\\
            \textbf{LSTM-4}   &0.0840&0.0849&0.0957&0.0289&0.0291&0.0309&0.0232&\textbf{0.0233}&0.0248&64.44&20.97&13.13\\
	    \textbf{Transformer-3} &0.1884&0.1678&0.2858&0.0434&0.0409&0.0534&0.0338&0.0323&0.0423&98.17&30.49&20.26\\
            \textbf{Transformer-4}   &0.1384&0.1875&0.3672&0.0372&0.0430&0.0659&0.0291&0.0335&0.0479&84.49&27.78&23.03\\
        \textbf{Mamba-3}&0.1055&0.0848&\textbf{0.0952}&0.0324&\textbf{0.0291}&\textbf{0.0308}&\textbf{0.0249}&0.0233&0.0244&\textbf{10.73}&\textbf{11.56}&9.50\\
	      \textbf{Mamba-4}&\textbf{0.0794}&\textbf{0.0827}&\textbf{0.0777}&0.0281&\textbf{0.0287}&\textbf{0.0278}&\textbf{0.0224}&0.0225&0.0219&\textbf{10.64}&\textbf{10.70}&9.87\\
	    \textbf{DualMamba-3}&\textbf{0.0765}&\textbf{0.1591}&\textbf{0.0987}&0.0276&\textbf{0.0398}&\textbf{0.0314}&\textbf{0.0224}&\textbf{0.0318}&\textbf{0.0244}&\textbf{8.79}&\textbf{12.54}&\textbf{8.08}\\
           \textbf{DualMamba-4}&\textbf{0.0663}&\textbf{0.0668}&\textbf{0.0853}&\textbf{0.0257}&\textbf{0.0258}&\textbf{0.0292}&\textbf{0.0196}&\textbf{0.0200}&\textbf{0.0225}&\textbf{7.60}&\textbf{10.95}&\textbf{6.32}\\
		\bottomrule
	\end{tabular}
\end{table*}

\section{Case Study and Result Analysis}
\label{sec:case-study}
\subsection{UK Climate Datasets}
To investigate the spatiotemporal variability of the UK climate, this study utilizes CET dataset, a high-resolution gridded observational dataset developed and maintained by the UK Met Office National Climate Information Centre. This dataset integrates observations from the UK land weather station network, adhering to World Meteorological Organization (WMO) best-practice standards, and undergoes rigorous quality assurance to ensure reliability for climate monitoring and trend analysis.

For this analysis, we extracted monthly mean temperature and precipitation data over the period 1961--2024, aligning with the WMO-recommended 30-year climatological baseline (1991--2020) and the longer-term reference period (1961--1990) to assess both contemporary anomalies and historical trends. The dataset's 1\, km resolution facilitates detailed regional analysis, while aggregated national and regional statistics provide a broader context for interpreting climate variability. 

These findings leverage the CET dataset’s high spatial fidelity and long temporal span, enabling robust detection of climate signals amidst natural variability. Uncertainties associated with station closures and data homogenization were minimized through the Met Office’s peer-reviewed interpolation techniques, ensuring the integrity of long-term trends. The dataset’s open accessibility via the Met Office Climate Data Portal further enhances its utility for reproducibility and interdisciplinary climate-related research. The data details and examples are given in Table~\ref{tab:cet_stations}.

\begin{table*}[h]
    \centering
    \renewcommand{\arraystretch}{0.9}
    \caption{DualMamba Hyperparameter Sensitivity Analysis on Testing Set.}
    \begin{tabular}{ccccccc}
    \toprule
       \textbf{Stations}  &  \textbf{Hyperparameter} & \textbf{MSE} & \textbf{RMSE} & \textbf{MAE} & \textbf{MAPE (\%)} & \textbf{Param.}\\
    \hline
\multirow{9}{*}{\textbf{Radcliffe}}            & L=1000, T=300  & 0.1194 & 0.3456 & 0.2816 & 10.31 & 157,596\\
                                               & L=1000, T=400  & 0.1197 & 0.3460 & 0.2818 & 10.46 & 159,296\\
                                               & L=1000, T=500  & 0.1196 & 0.3459 & 0.2817 & 10.25 & 160,996\\
                                               & L=800, T=300   & 0.1198 & 0.3462 & 0.2820 & 10.77 & 144,796\\
                                               & L=800, T=400   & 0.1201 & 0.3466 & 0.2823 & 10.88 & 146,496\\
                                               & L=800, T=500   & 0.1196 & 0.3459 & 0.2817 & 10.35 & 148,196\\
                                               & L=600, T=300   & 0.1195 & 0.3456 & 0.2816 & 10.44 & 131,996\\
                                               & L=600, T=400   & 0.1196 & 0.3458 & 0.2817 & 10.39 & 133,696\\
                                               & L=600, T=500   & 0.1198 & 0.3461 & 0.2819 & 10.18 & 135,396\\
    \hline
\multirow{9}{*}{\textbf{Rothamsted}}           & L=1000, T=300  & 0.1154 & 0.3396 & 0.2779 & 10.42 & 157,596\\
                                               & L=1000, T=400  & 0.1158 & 0.3403 & 0.2784 & 10.35 & 159,296\\
                                               & L=1000, T=500  & 0.1157 & 0.3402 & 0.2784 & 10.35 & 160,996\\
                                               & L=800, T=300   & 0.1157 & 0.3402 & 0.2782 & 10.77 & 144,796\\
                                               & L=800, T=400   & 0.1151 & 0.3393 & 0.2776 & 10.38 & 146,496\\
                                               & L=800, T=500   & 0.1159 & 0.3405 & 0.2786 & 10.36 & 148,196\\
                                               & L=600, T=300   & 0.1157 & 0.3401 & 0.2782 & 10.38 & 131,996\\
                                               & L=600, T=400   & 0.1153 & 0.3396 & 0.2778 & 10.31 & 133,696\\
                                               & L=600, T=500   & 0.1156 & 0.3399 & 0.2781 & 10.19 & 135,396\\
    \hline
\multirow{9}{*}{\textbf{Ross\_on\_Wye}}        & L=1000, T=300  & 0.1264 & 0.3555 & 0.2928 & 26.76 & 157,596\\
                                               & L=1000, T=400  & 0.1265 & 0.3557 & 0.2930 & 19.44 & 159,296\\
                                               & L=1000, T=500  & 0.1270 & 0.3563 & 0.2935 & 17.38 & 160,996\\
                                               & L=800, T=300   & 0.1283 & 0.3582 & 0.2951 & 70.58 & 144,796\\
                                               & L=800, T=400   & 0.1269 & 0.3562 & 0.2934 & 17.07 & 146,496\\
                                               & L=800, T=500   & 0.1271 & 0.3565 & 0.2937 & 24.21 & 148,196\\
                                               & L=600, T=300   & 0.1263 & 0.3554 & 0.2927 & 37.08 & 131,996\\
                                               & L=600, T=400   & 0.1263 & 0.3554 & 0.2927 & 15.44 & 133,696\\
                                               & L=600, T=500   & 0.1269 & 0.3562 & 0.2933 & 18.69 & 135,396\\
    \hline
\multirow{9}{*}{\textbf{Stonghurst}}           & L=1000, T=300  & 0.1141 & 0.3378 & 0.2766 & 10.24 & 157,596\\
                                               & L=1000, T=400  & 0.1140 & 0.3377 & 0.2766 & 10.14 & 159,296\\
                                               & L=1000, T=500  & 0.1140 & 0.3376 & 0.2765 & 10.20 & 160,996\\
                                               & L=800, T=300   & 0.1198 & 0.3462 & 0.2820 & 10.28 & 144,796\\
                                               & L=800, T=400   & 0.1143 & 0.3380 & 0.2768 & 10.34 & 146,496\\
                                               & L=800, T=500   & 0.1144 & 0.3382 & 0.2770 & 10.38 & 148,196\\
                                               & L=600, T=300   & 0.1141 & 0.3377 & 0.2767 & 10.68 & 131,996\\
                                               & L=600, T=400   & 0.1140 & 0.3376 & 0.2765 & 10.64 & 133,696\\
                                               & L=600, T=500   & 0.1139 & 0.3375 & 0.2765 & 10.16 & 135,396\\
    \bottomrule
    \end{tabular}
    \label{tab5}
\end{table*}
\begin{table*}[h]
    \centering
    \renewcommand{\arraystretch}{0.9}
    \caption{DualMamba Hyperparameter Sensitivity Analysis.}
    \begin{tabular}{ccccccc}
    \toprule
       \textbf{Stations}  &  \textbf{Hyperparameter} & \textbf{MSE} & \textbf{RMSE} & \textbf{MAE} & \textbf{MAPE (\%)} & \textbf{Param.}\\
    \hline
\multirow{9}{*}{\textbf{Malvern}}              & L=1000, T=300  & 0.1244 & 0.3527 & 0.2867 & 17.26 & 157,596\\
                                               & L=1000, T=400  & 0.1247 & 0.3531 & 0.2871 & 18.99 & 159,296\\
                                               & L=1000, T=500  & 0.1242 & 0.3524 & 0.2866 & 20.79 & 160,996\\
                                               & L=800, T=300   & 0.1242 & 0.3525 & 0.2866 & 13.74 & 144,796\\
                                               & L=800, T=400   & 0.1241 & 0.3523 & 0.2865 & 19.02 & 146,496\\
                                               & L=800, T=500   & 0.1246 & 0.3530 & 0.2872 & 24.24 & 148,196\\
                                               & L=600, T=300   & 0.1246 & 0.3529 & 0.2870 & 29.55 & 131,996\\
                                               & L=600, T=400   & 0.1243 & 0.3525 & 0.2868 & 12.13 & 133,696\\
                                               & L=600, T=500   & 0.1245 & 0.3529 & 0.2872 & 16.36 & 135,396\\
    \hline
\multirow{9}{*}{\textbf{Squires\_Gate}}        & L=1000, T=300  & 0.1043 & 0.3230 & 0.2625 & 10.27 & 157,596\\
                                               & L=1000, T=400  & 0.1042 & 0.3228 & 0.2623 & 10.38 & 159,296\\
                                               & L=1000, T=500  & 0.1046 & 0.3234 & 0.2630 & 10.37 & 160,996\\
                                               & L=800, T=300   & 0.1042 & 0.3228 & 0.2622 & 10.37 & 144,796\\
                                               & L=800, T=400   & 0.1043 & 0.3230 & 0.2625 & 10.81 & 146,496\\
                                               & L=800, T=500   & 0.1043 & 0.3229 & 0.2626 & 10.17 & 148,196\\
                                               & L=600, T=300   & 0.1038 & 0.3222 & 0.2617 & 10.42 & 131,996\\
                                               & L=600, T=400   & 0.1039 & 0.3223 & 0.2619 & 10.21 & 133,696\\
                                               & L=600, T=500   & 0.1051 & 0.3242 & 0.2635 & 10.93 & 135,396\\
    \hline
\multirow{9}{*}{\textbf{Ringway}}              & L=1000, T=300  & 0.0930 & 0.3050 & 0.2543 & 38.15 & 157,596\\
                                               & L=1000, T=400  & 0.0932 & 0.3053 & 0.2546 & 22.21 & 159,296\\
                                               & L=1000, T=500  & 0.0933 & 0.3055 & 0.2547 & 12.07 & 160,996\\
                                               & L=800, T=300   & 0.0930 & 0.3049 & 0.2543 & 18.00 & 144,796\\
                                               & L=800, T=400   & 0.0933 & 0.3054 & 0.2547 & 11.29 & 146,496\\
                                               & L=800, T=500   & 0.0931 & 0.3052 & 0.2543 & 14.88 & 148,196\\
                                               & L=600, T=300   & 0.0932 & 0.3054 & 0.2547 & 21.43 & 131,996\\
                                               & L=600, T=400   & 0.0932 & 0.3053 & 0.2545 & 24.34 & 133,696\\
                                               & L=600, T=500   & 0.0931 & 0.3052 & 0.2543 & 10.08 & 135,396\\
    \hline
\multirow{9}{*}{\textbf{Pershore\_College}}    & L=1000, T=300  & 0.1401 & 0.3743 & 0.3071 & 27.63 & 157,596\\
                                               & L=1000, T=400  & 0.1390 & 0.3728 & 0.3054 & 25.31 & 159,296\\
                                               & L=1000, T=500  & 0.1390 & 0.3728 & 0.3055 & 16.83 & 160,996\\
                                               & L=800, T=300   & 0.1391 & 0.3729 & 0.3059 & 28.84 & 144,796\\
                                               & L=800, T=400   & 0.1386 & 0.3723 & 0.3051 & 16.00 & 146,496\\
                                               & L=800, T=500   & 0.1390 & 0.3728 & 0.3055 & 20.50 & 148,196\\
                                               & L=600, T=300   & 0.1395 & 0.3735 & 0.3064 & 44.17 & 131,996\\
                                               & L=600, T=400   & 0.1391 & 0.3730 & 0.3057 & 21.75 & 133,696\\
                                               & L=600, T=500   & 0.1391 & 0.3730 & 0.3059 & 24.01 & 135,396\\
    \bottomrule
    \end{tabular}
    \label{tab6}
\end{table*}

 \subsection{Model Implementation Details and Evaluation Metrics}
\label{subsec:model_impl_eval}
 The model was trained and evaluated using the CET dataset, which provides gridded monthly temperature data at 1\,km $\times$ 1\, km resolution from 1961 to 2024. To optimize computational efficiency while preserving climatic fidelity, our methodology employs z-score normalized national monthly temperature anomalies partitioned through stratified temporal blocking: 70\% training, 15\% validation and 15\% test.  

The DualMamba architecture comprised an input layer with 1000 neurons, two hidden layers each with 64 neurons, and 300 output neuron for the predicted temperature. Unlike traditional neural networks, DualMamba uses bidirectional Mamba encoders to capture the intricate spatiotemporal dynamics of UK climate data. Traditional RNN struggle to capture long-term dependencies in sequences due to issues like gradient varnishing and explosion. Although variants like LSTM and GRU have addressed these issues to some extent, they still face challenges with very long sequences. In our experimental design, RNN, GRU, LSTM, Transformer, Mamba-villa are also selected as the baselines. The detailed model architecture is presented in Table \ref{tab:architecture}. The Mamba layer is set as $L=2$, and adaptive residual coefficient is set as 0.5. The model was trained using the AdamW optimizer with the learning rate 0.0001 over 50 epochs with a batch size of 4096, minimizing the mean squared error (MSE) loss. The dropout rate is set as 0.1. Validation set performance was monitored to implement early stopping with 10 patience, ensuring the selection of the best-performing model for testing.  The proposed DualMamba is running on a high performance computation with PyTorch framework and with 2 Nvidia H200s graph card with 282GB memory for the accelerated computation. The CUDA 12.8 is used for the tensor computation. In addition, distributed data parallel techniques is used for the computation acceleration of training. In this way, we speedup the training process, DualMamba’s inference latency can be reduced to within $1.1\times $ that of an MLP by precomputing and fusing SiLU activation into optimized CUDA cache kernels.

Model performance was evaluated using four metrics: MSE, RMSE, MAE, and MAPE, computed on the original temperature scale for evaluating the performance. Their mathematical definitions are provided below. MSE quantifies the average squared deviation, emphasizing larger errors:
\begin{equation}
    \text{MSE} = \frac{1}{n} \sum_{i=1}^{n} (y_i - \hat{y}_i)^2.
    \label{mse}
\end{equation}
In the prediction tasks, it highlights variance in predictive accuracy. RMSE measures error in the same units as temperature, balancing sensitivity to outliers:
\begin{equation}
\text{RMSE} = \sqrt{\frac{1}{n} \sum_{i=1}^{n} (y_i - \hat{y}_i)^2}.
\label{rmse}
\end{equation}
RMSE provides an intuitive error magnitude. MAE assesses the average absolute deviation, robust to extreme errors:
\begin{equation}
    \text{MAE} = \frac{1}{n} \sum_{i=1}^{n} |y_i - \hat{y}_i|.
    \label{mae}
\end{equation}
MAE is a critical index for evaluating the typical prediction error. MAPE expresses error relative to observed values, enabling scale-independent comparison:
\begin{equation}
        \text{MAPE} = \frac{1}{n} \sum_{i=1}^{n} \left| \frac{y_i - \hat{y}_i}{y_i} \right| \times 100,
        \label{mape}
\end{equation}
where $y_i$ is the observed temperature, $\hat{y}_i$ the predicted value, and $n$ the number of test samples. It is known that MAPE is particularly suited for variables with varying ranges.

These metrics were computed on the test set, with results benchmarked against baseline model Mamba to assess DualMamba's efficacy. Hyperparameter tuning, including different input interval, was guided by validation performance, optimizing for RMSE. Detailed results and regional comparisons are presented in Section~\ref{sec4}, demonstrating DualMamba's potential as a robust tool for climate prediction.
\begin{figure*}[]
    \centering
    \includegraphics[width=0.8\linewidth]{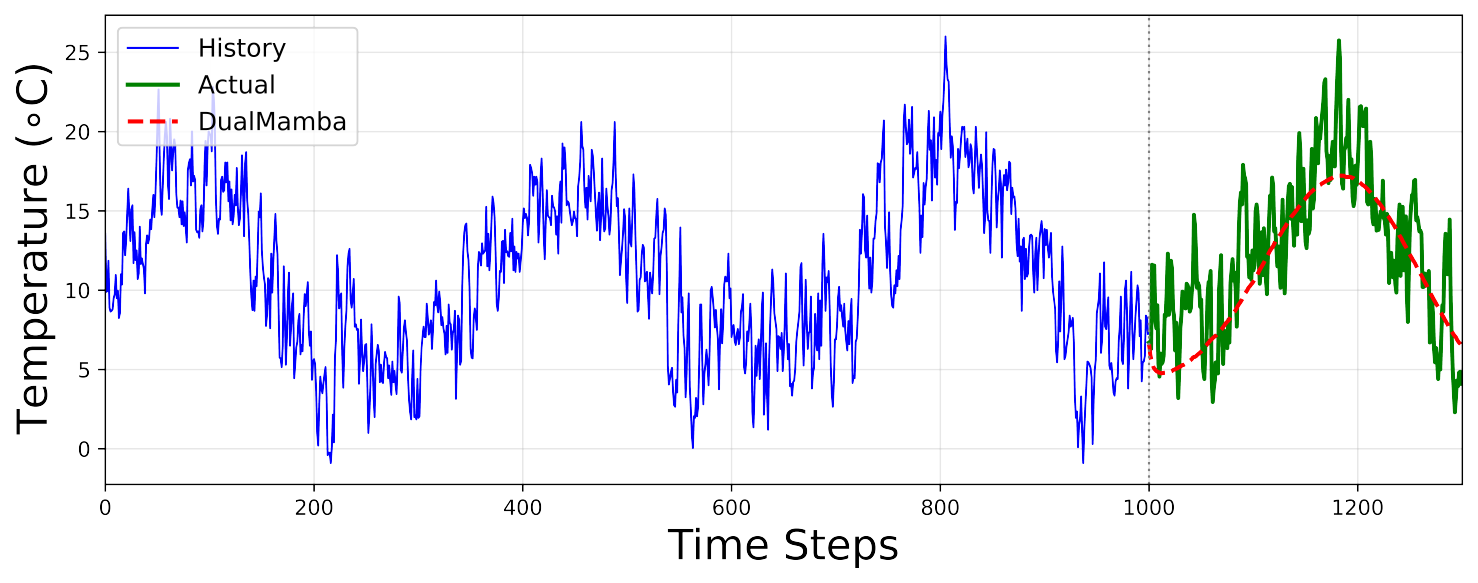}
    \hfil
    \includegraphics[width=0.8\linewidth]{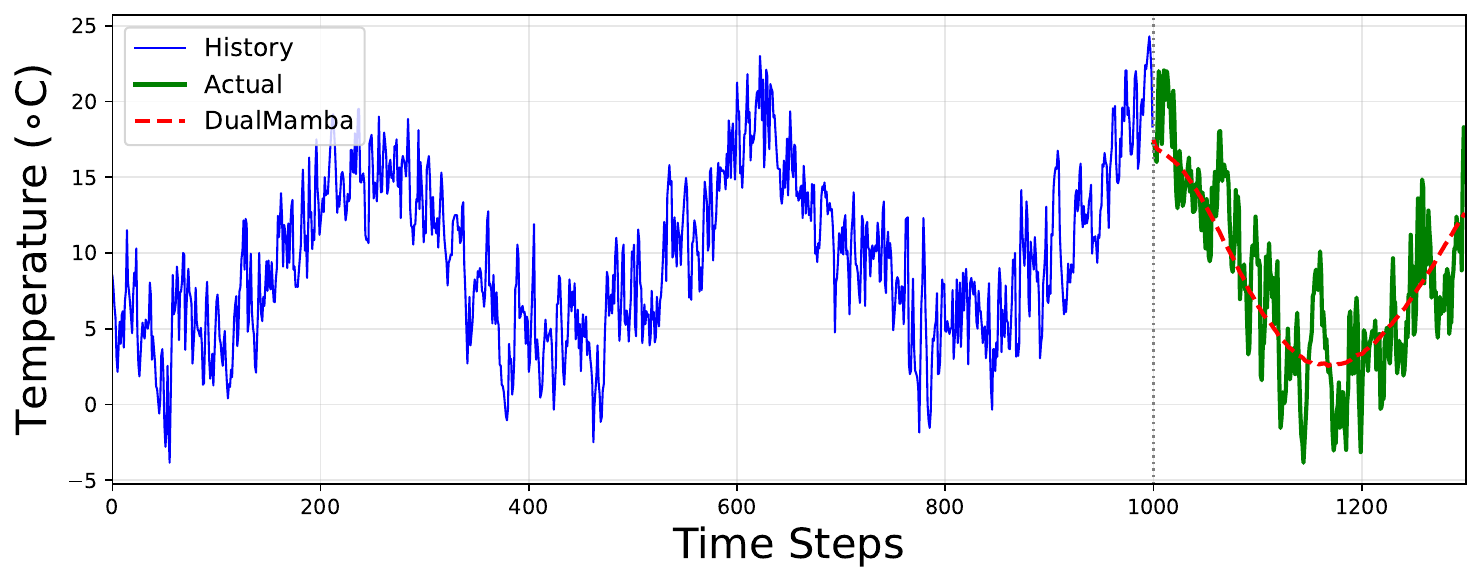}
    \hfil
    \includegraphics[width=0.8\linewidth]{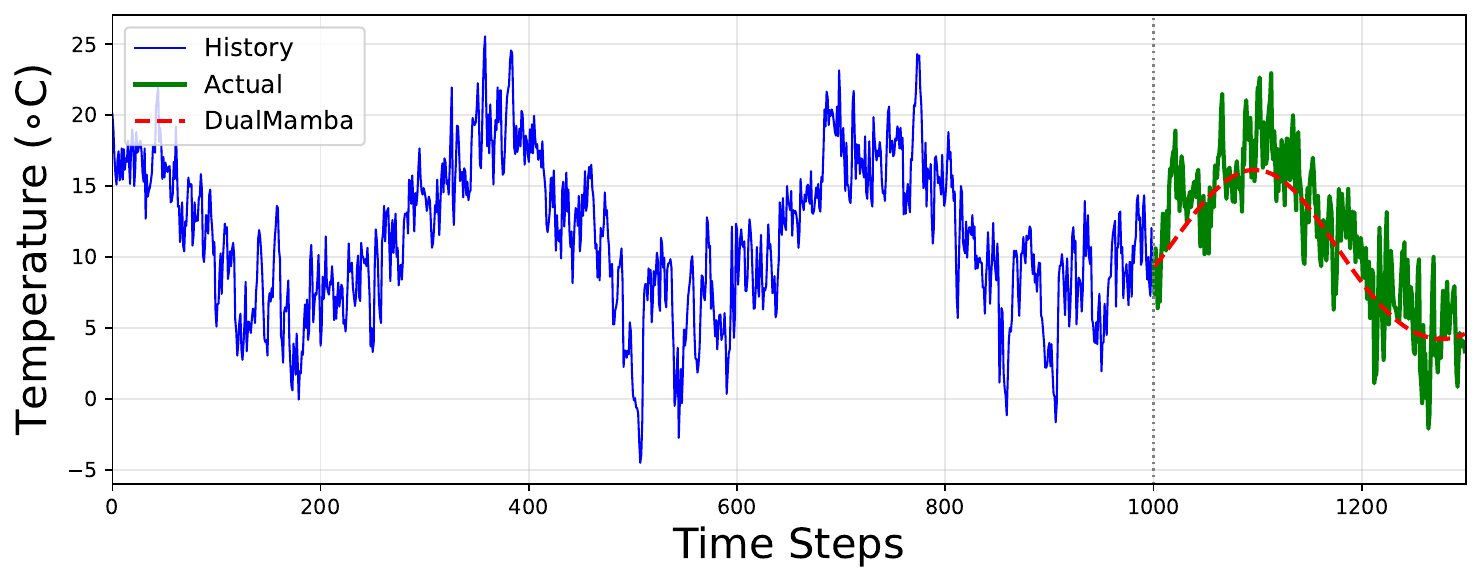}
    \hfil
    \includegraphics[width=0.8\linewidth]{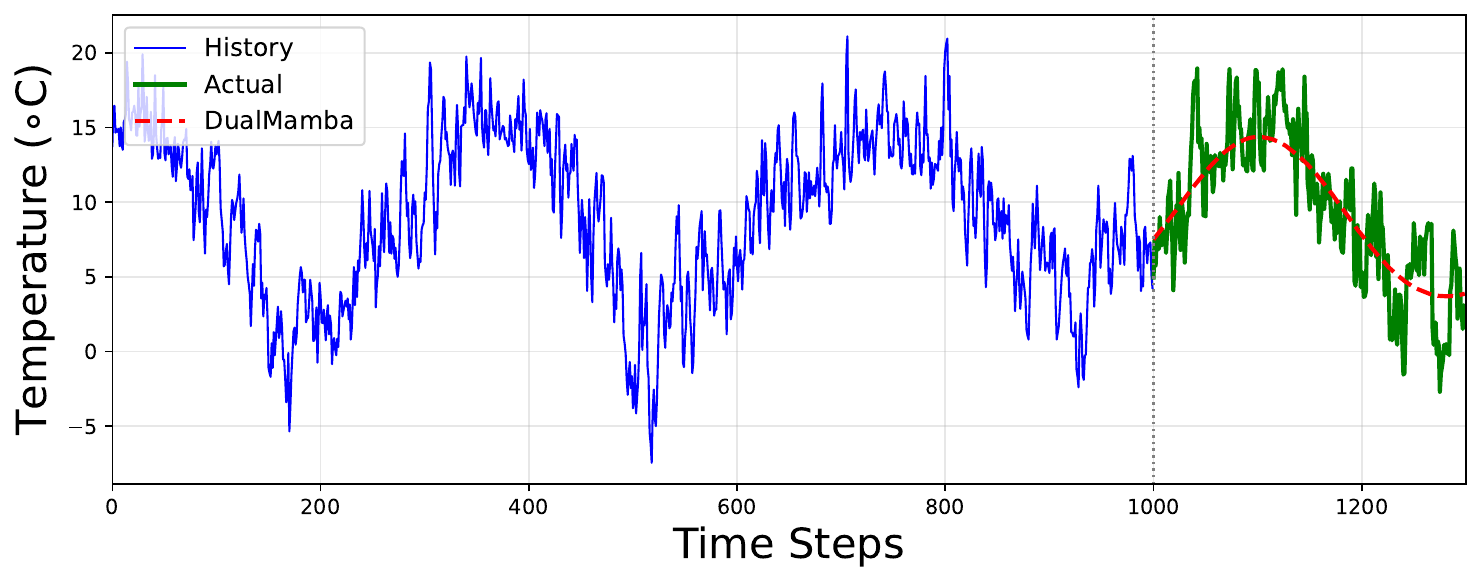}
    \caption{The Historical Observation and Prediction Comparisons with Radcliffe, Rothamsted, Ross\_on\_Wye, and Stonyhurst, Respectively.}
    \label{fig:results1}
\end{figure*}

\begin{figure*}[]
    \centering
    \includegraphics[width=0.8\linewidth]{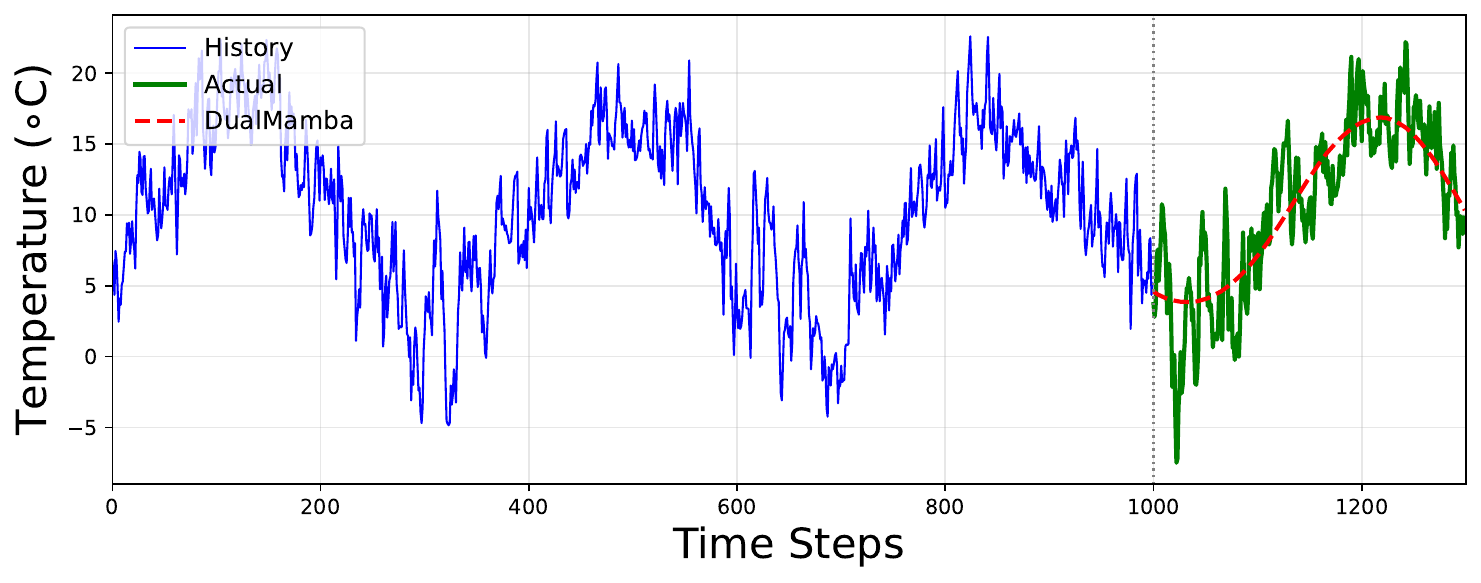}
    \hfil
    \includegraphics[width=0.8\linewidth]{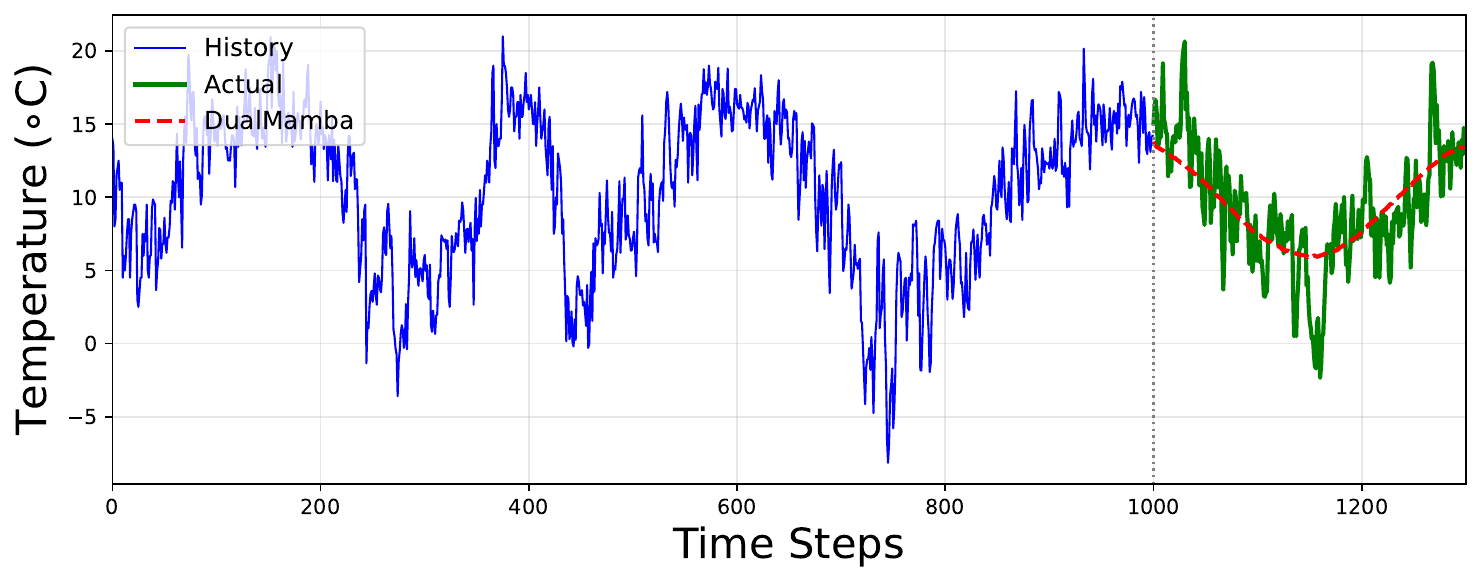}
    \hfil
    \includegraphics[width=0.8\linewidth]{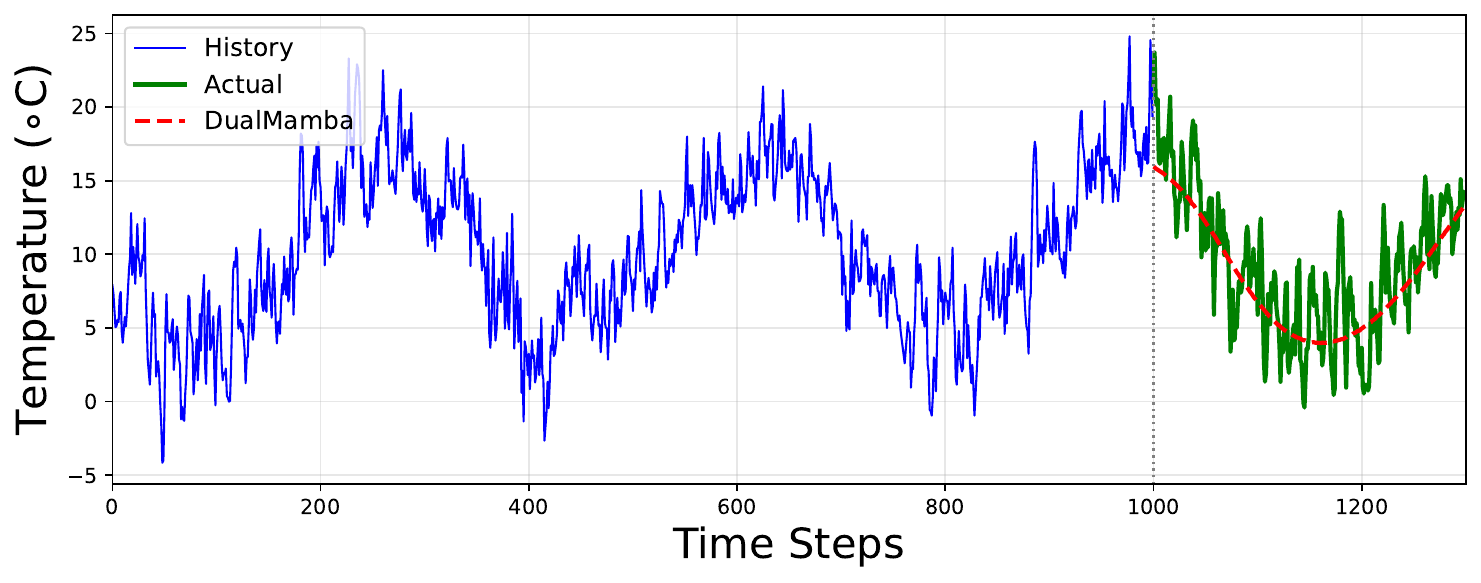}
    \hfil
    \includegraphics[width=0.8\linewidth]{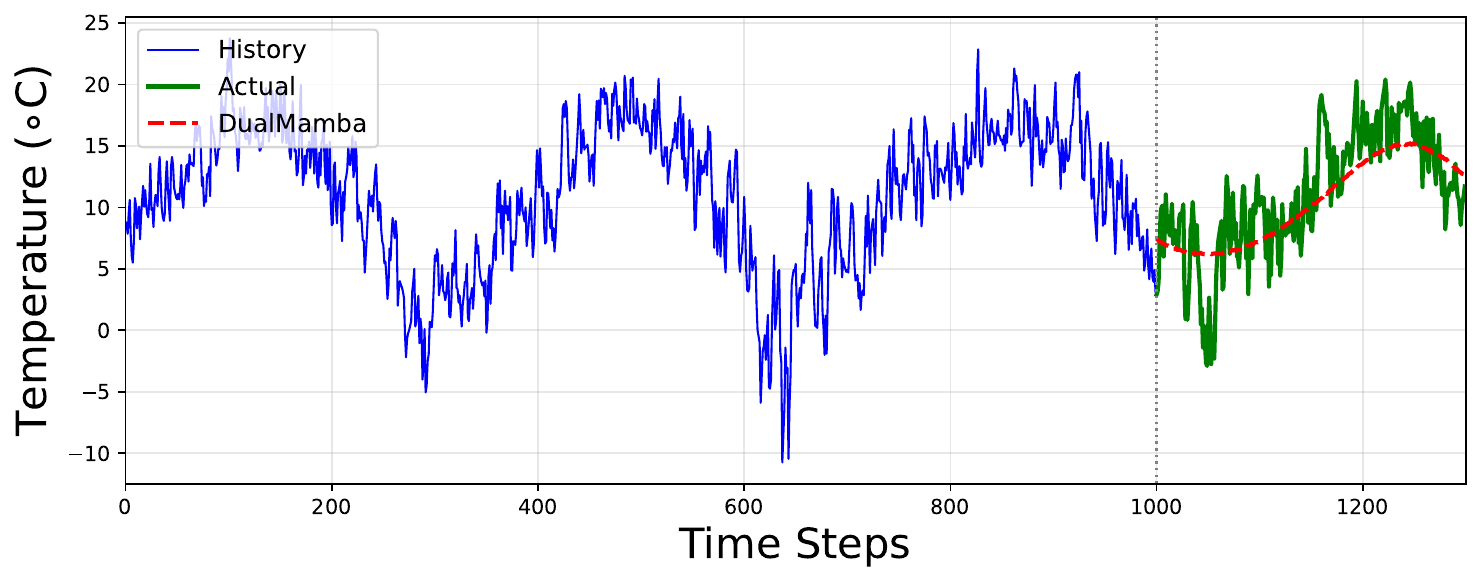}
    \caption{The Historical Observation and Prediction Comparisons with Malvern, Squires\_Gate, Ringway, and Pershore\_College, Respectively.}
    \label{fig:results2}
\end{figure*}

\begin{figure*}[h]
    \centering
    \includegraphics[width=0.8\linewidth]{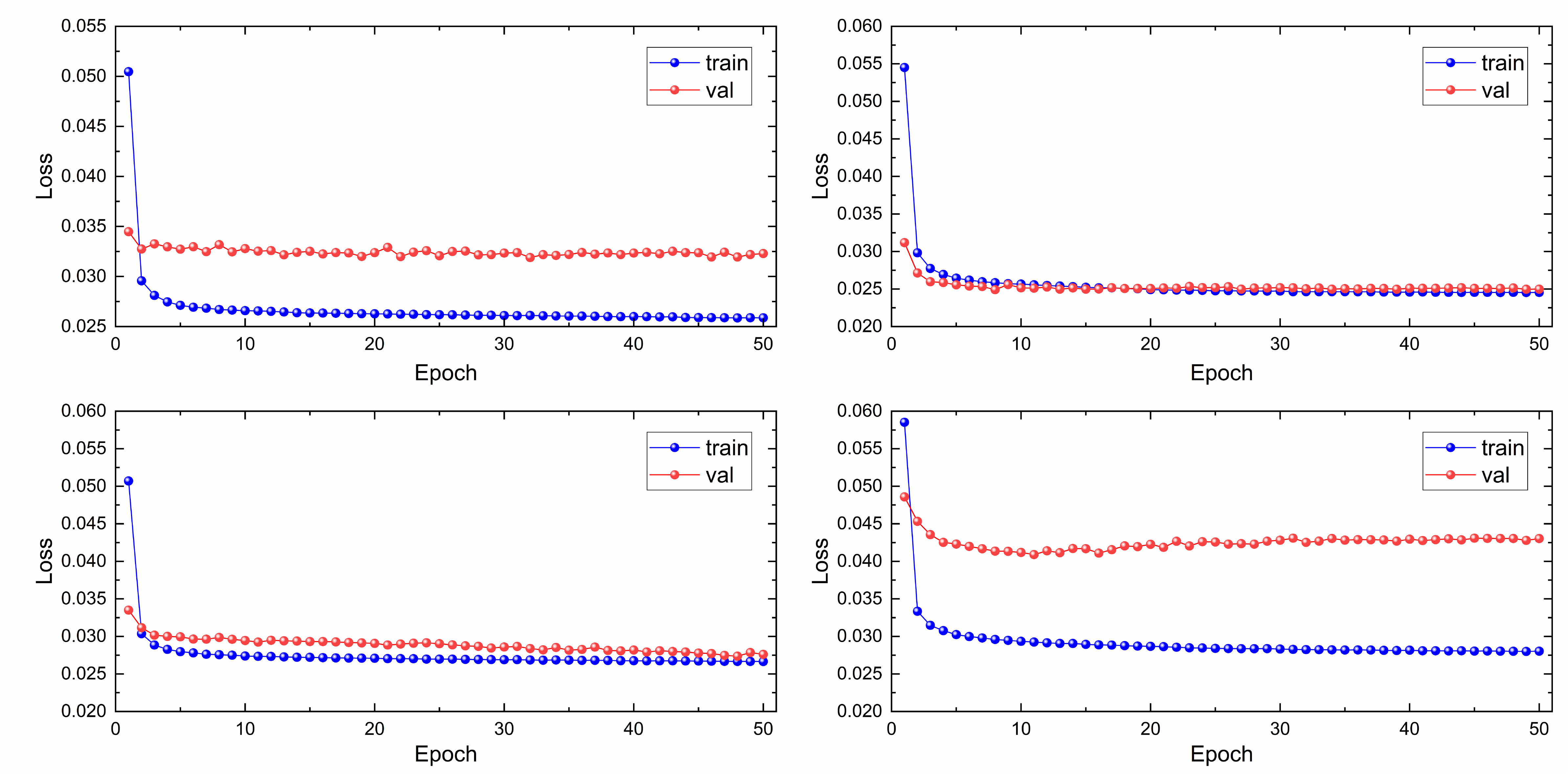}
    \includegraphics[width=0.8\linewidth]{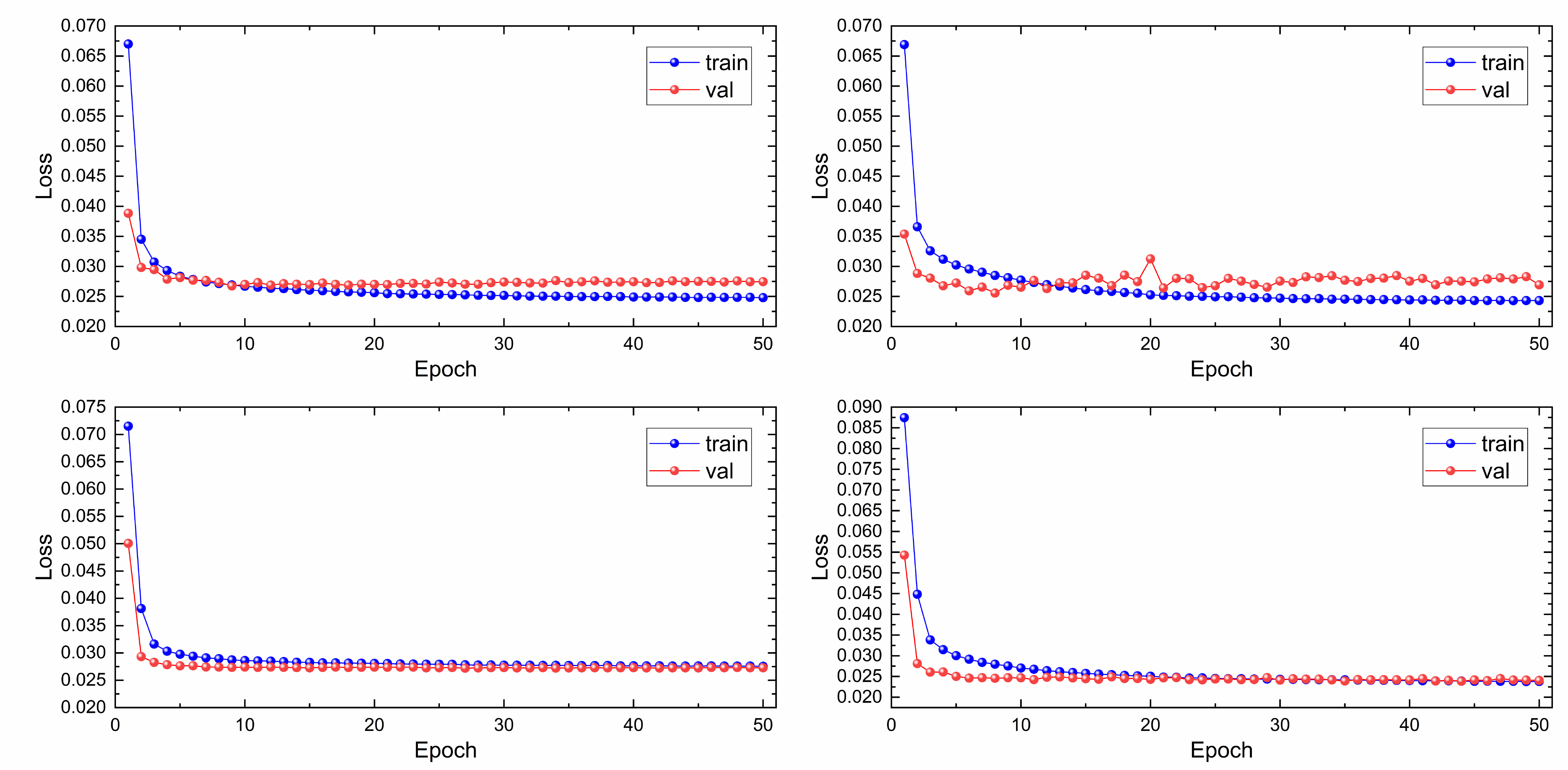}
    \caption{Training and Validation Loss on Radcliffe, Ross\_on\_Wye, Rothamsted, Stonyhurst, Malvern, Squires Gate, Ringway, and Pershore College, Respectively.}
    \label{fig:RRRS}
\end{figure*}

\subsection{Simulation Result Analysis}
In our experimental design, we compare MLP-3, MLP-4, RNN-3, RNN-4, GRU-3, GRU-4, LSTM-3, LSTM-4, Transformer-3, Transformer-4, and Mamba-3, Mamba-4, respectively. In addition, the {\bf bold} font in the table indicates superior results. Table \ref{presult} summarizes all the comparison results for different approaches and our presented DualMamba. It shows that our presented DualMamba outperforms RNN and LSTM in the four indexes, which means DualMamba achieves higher prediction accuracy (lower error index). It is worth noting that, compared to Transformer, our presented DualMamba have a superior forecasting performance. Furthermore, compared to villa-Mamba, our proposed DualMamba also achieves better forecasting performance, and an intuitive lower error indexes, which means that DualMamba learns better features representation by using bidirectional encoders. 

Figure \ref{fig:radar_charts} gives the radar chart which provides a comprehensive comparison of six distinct models, namely MLP-3, RNN-3, GRU-3, LSTM-3, Transformer-3, and DualMamba-3, across four key performance metrics. Each axis on the radar chart corresponds to one of the datasets: Training, Validation, and Testing. The data points for each model are connected by a line, with the solid line representing the DualMamba-3 model and the dashed lines representing the other models for clear distinction. Notably, the DualMamba-3 model is highlighted by a solid line and filled area, emphasizing its comparative performance against the other models represented by dashed lines without fill.
The chart effectively visualizes the balanced performance of the DualMamba-3 model across different datasets, particularly excelling in the Training dataset as indicated by its proximity to the chart's center. In contrast, the Transformer-3 model shows a notable deviation in the Testing dataset, suggesting potential overfitting or dataset-specific performance issues. This visualization aids in quickly assessing the robustness and generalizability of each model under evaluation.

Figure \ref{fig:results1} and Figure \ref{fig:results2} quantifies the forecasting fidelity of DualMamba against conventional Mamba on the eight stations testing benchmark. The results demonstrate DualMamba's superior capacity to resolve multiscale temperature dynamics, achieving a 0.003 reduction in RMSE. Compared to MLPs, DualMamba captures better local dynamic informatics, while MLP only predicts the trends. This enhanced resolution stems from DualMamba's spline-parameterized architecture, which adaptively allocates basis functions to critical dynamic regimes. Figure \ref{fig:RRRS} describes the corresponding training process for DualMamba, it is observable that all the eights UK dataset coverages to a stable value, which also matches our theoretical convergence analysis. The training process is stable with all the eight benchmarks even in Radcliffe and Stonyhurst, the train and validation loss has a permitable margin. 

To conduct a comprehensive ablation study, we vary both the historical horizon and the prediction horizon. Specifically, the historical horizons are set to [1000, 800, 600], while the prediction horizons are [300, 400, 500], respectively. Table \ref{tab5} presents a systematic hyperparameter sensitivity analysis of the DualMamba framework. Due to the similar results on other seven stations, we only shows the Radcliffe comparisons, other seven results are omitted. An intuitive result from Table \ref{tab5} is that the DualMamba achieves extreme superior performance especially in terms of both long-term forecasting and forecasting robustness with different forecasting horizons. From historical horizon 1000 to 600, the decreasing horizon would not causes a performance loss, even with a large prediction range. In the worst forecasting case, 600 historical horizon to predict 500 preceding horizon, we still have the 0.1198 RMSE, which is similar to the best case, 1000 historical horizon to predict 300 horizon. Other forecasting approaches performance largely depends on the historical length, the performance decrease largely when decreasing the historical interval, or increasing the forecasting horizon. 
\section{Conclusion}
\label{sec5}
Climate temperature in the UK is critical for monitoring carbon emissions, improving the goal of global warming. In this brief, an improved DualMamba framework is proposed for temperature prediction. It shows that DualMamba improves the prediction accuracy compared to the traditional MLP. The main advantages of the presented DualMamba are:

1) A novel adaptive deep DualMamba architecture for UK temperature prediction is presented.

2) The flexibility and hyper-parameters sensitivity of the proposed DualMamba has been evaluated in the ablation study.

3) Rigorous convergence analysis establishing the universal approximation property. It also have the advantage of mathematical guarantee
\[\lim_{\epsilon \to 0} \frac{\mathcal{N}_{\mathrm{MLP}}(\epsilon)}{\mathcal{N}_{\mathrm{DualMamba}}(\epsilon)} = \infty.\] It is superior performance improvement in long-term prediction.

4) The results indicate that DualMamba requires low computational overhead and achieves leading robust performance. It achieves extraordinary in long-term time-series forecasting robustness.

The advantage is primarily attributed to the bidirectional Mamba block within the dual encoding Layer, which offers an enhanced understanding of climate temperature at a lower overhead compared to the Transformer. 
Furthermore, we conduct extensive experiments to prove Mamba possesses robust capabilities in time-series forecasting tasks. We demonstrate that DualMamba has the forecasting performance in long-term prediction with acceptable hyperparameter.

Notwithstanding these advancements, the current DualMamba implementation exhibits prolonged training durations due to long dependency. Mamba excels at detecting long-range dependencies and controlling the escalation of computational demands, thus equipping it to meet the challenges outlined. In climate temperature forecasting tasks where resources are limited, compared with Transformer, Mamba is capable of processing more variates information simultaneously and delivering more accurate predictions. The proposed DualMamba is a general forecasting framework which can be extended to other time-series forecasting domains. Looking forward, Mamba-based models are expected to be applicable to a broader spectrum of time series prediction tasks that involve processing extensive variate data. 

Pretrained models utilizing the Transformer architecture capitalize on its robust generalization capabilities, achieving notable success in climate temperature prediction. Our experimental results indicate that Mamba matches the Transformer in both generalization and stability, suggesting that the development of a Mamba-based pre-training model for other time-series forecasting tasks could be a fruitful direction to explore. This work bridges theoretical approximation theory with practical climate informatics, providing a novel paradigm for high-dimensional spatiotemporal prediction tasks.

\bibliographystyle{IEEEtran}
\bibliography{reference}

\vfill

\end{document}